\documentclass[sn-mathphys,Numbered]{sn-jnl}

\usepackage{graphicx}%
\usepackage{multirow}%
\usepackage{amsmath,amssymb,amsfonts}%
\usepackage{amsthm}%
\usepackage{mathrsfs}%
\usepackage[title]{appendix}%
\usepackage{xcolor}%
\usepackage{textcomp}%
\usepackage{manyfoot}%
\usepackage{booktabs}%
\usepackage{algorithm}%
\usepackage{algorithmicx}%
\usepackage{algpseudocode}%
\usepackage{listings}%
\usepackage{longtable}%

\theoremstyle{thmstyleone}%
\theoremstyle{thmstyletwo}%

\theoremstyle{thmstylethree}%

\begin{document}

\title[Article Title] {Present and Future of AI in Renewable Energy Domain : A Comprehensive Survey}


\author[1]{\fnm{Abdur} \sur{Rashid}}\email{rabdurrashid091@gmail.com}

\author[2]{\fnm{Parag} \sur{Biswas}}\email{text2parag@gmail.com}

\author[3]{\fnm{Angona} \sur{Biswas}}\email{angonabiswas28@gmail.com}

\author*[4]{\fnm{MD Abdullah Al} \sur{Nasim}}\email{nasim.abdullah@ieee.org}

\author[5]{\fnm{Kishor} \sur{Datta Gupta}}\email{kgupta@cau.edu}

\author[6]{\fnm{Roy} \sur{George}}\email{george@cau.edu}

\affil[1, 2]{\orgdiv{MSEM Department}, \orgname{Westcliff university},  \orgaddress{{\city{California}, \country{United States}}}}

\affil[3, 4]{\orgdiv{Research and Development Department}, \orgname{Pioneer Alpha},  \orgaddress{{\city{Dhaka}, \country{Bangladesh}}}}

\affil[5, 6]{\orgdiv{Department of Computer and Information Science}, \orgname{Clark Atlanta University}, {\city{Georgia}, \country{USA}}}

\abstract{Artificial intelligence (AI) is becoming an important tool for improving processes in many industries, including electrical power systems, thanks to advancements in digital technology.  This paper starts by reviewing how AI is currently used in renewable energy (RE) applications. It then provides an in-depth look at renewable energy systems, assessing their effectiveness and identifying the most commonly used AI algorithms. The review highlights nine AI-based methods that help integrate renewable energy into modern power systems. It also gives a clear overview of AI techniques used in different areas of renewable energy. By analyzing several studies, the paper offers useful insights into the strengths and weaknesses of these methods. Additionally, it focuses on three main areas: how AI is used to generate renewable energy, forecast energy production, and optimize energy systems. The paper also shows how AI outperforms traditional methods in areas like robotics, preventing cyberattacks, processing data, and managing smart grids. The findings show that AI has great potential to improve the energy sector, making it more efficient, secure, and adaptable. The paper concludes by discussing future directions for using AI in renewable energy, highlighting its role in shaping the future of sustainable energy systems.}

\keywords{Artificial Intelligence (AI), renewable energy, Machine Learning, Deep Learning, Power Consumption, Power System}



\maketitle

\section{Introduction}\label{sec1}

The fundamental component of financially sound, ecologically responsible, and sustainable electricity generation is renewable energy, or RE. The role that renewable energy plays in reducing climate change \cite{he2023role}, \cite{lorente2023dynamic}, \cite{anser2024formulating} is among the most compelling arguments for adopting it. Compared to fossil fuels, renewable energy sources such as hydroelectric, solar, and wind power produce minimal to no greenhouse gas emissions throughout the energy generation process \cite{al2023impacts}. Renewable energy sources outperform finite fossil fuels in terms of sustainability and security \cite{yi2023environmental}. Natural replenishment of renewable resources, such as sunshine, wind, and water, makes them a steady and dependable energy source free from the price volatility and geopolitical unpredictability of fossil fuels \cite{laskaratou2023energy}. By supplying power to isolated and neglected areas, renewable energy technologies provide a viable way to combat energy poverty \cite{bamisile2023geothermal}. Off-grid solar systems may provide rural communities with dependable and reasonably priced power, raising living standards and promoting economic growth. Not only is renewable energy crucial, but it is also necessary for a thriving and sustainable future. We can solve a number of issues, including as public health, economic growth, energy security, and climate change, while building a more resilient and sustainable environment for coming generations by embracing renewable energy sources and moving toward a clean energy economy \cite{qiu2023energy}, \cite{li2023energy}.

\begin{figure}[H]
\includegraphics[width= 11 cm]{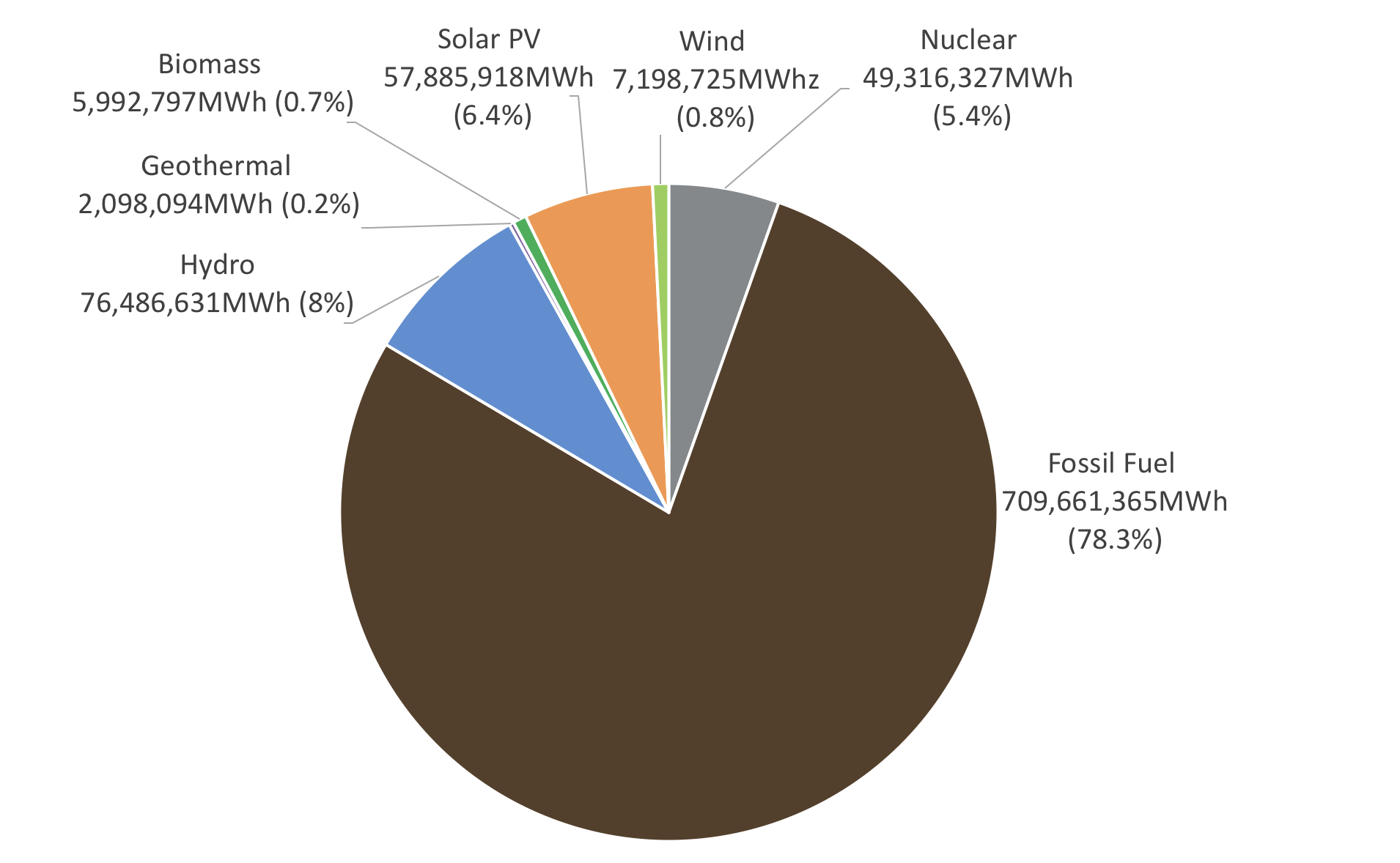}
\centering
\caption{The data in the pie chart shows the amount of energy that Japan's power transmission and distribution networks transported in 2018. \cite{renewable_energy_ratio_2018}
\label{fig1}}
\end{figure}

In accordance with the growing adoption of renewable energy sources to fulfill global energy demands, the International Energy Agency's (IEA) official report states that from 2019 \cite{ang2022comprehensive}, there has been a decrease in the requirement for fossil fuels to come up with power. RE technology research is always growing in an attempt to increase energy conversion efficiency, especially in terms of generation. The contemporary worldwide demand, especially in developed and emerging countries, demands the substitution of environmentally friendly electricity technology for conventional electrical generating resources including fossil fuels \cite{rizzi2014production}. Global warming and change in the climate are two harmful environmental problems brought on by fossil fuel-based energy sources. The amount of greenhouse gases that electricity generation has released into the atmosphere has increased exponentially during the last several decades. Renewable energy (RE) technologies, such as wind, hydroelectricity, solar energy, biomass, geothermal energy, and hydrogen energies, have been utilized to generate power in order to address the present environmental crisis \cite{ludin2018prospects}. It is projected that 100\% of the electricity required to generate electricity residential areas will come from renewable energy sources in the future. Data regarding Japan's 2018 renewable energy percentage is available from the Institute for Sustainable Energy Policies (ISEP) is shown in Figure \ref{fig1}. 

The crucial data on renewable energy is provided by the International Renewable Energy Agency (IRENA) as well as the Renewable Energy Policy Network over the 21st Century (REN21) \cite{IRENA-REN21}, and is included in Table \ref{renewable_capacity}. 

\begin{table}[htbp]
  \centering
  \caption{Global Renewable Energy Capacity by Source (2020-2023)}
  \label{tab:renewable_capacity}
  \begin{tabular}{lccccc}
    \toprule
    \textbf{Year} & \textbf{Hydro (GW)} & \textbf{Wind (GW)} & \textbf{Solar (GW)} & \textbf{Bioenergy (GW)} & \textbf{Geothermal (GW)} \\
    \midrule
    2020 & 1305 & 732 & 707 & 121 & 14 \\
    2021 & 1300 & 780 & 900 & 130 & 13 \\
    2022 & 1350 & 850 & 1000 & 140 & 15 \\
    2023 & 1400 & 900 & 1100 & 150 & 16 \\
    \bottomrule
  \end{tabular}
  \label{renewable_capacity}
\end{table}

\unskip

\begin{figure}[H]
\includegraphics[width= 11 cm]{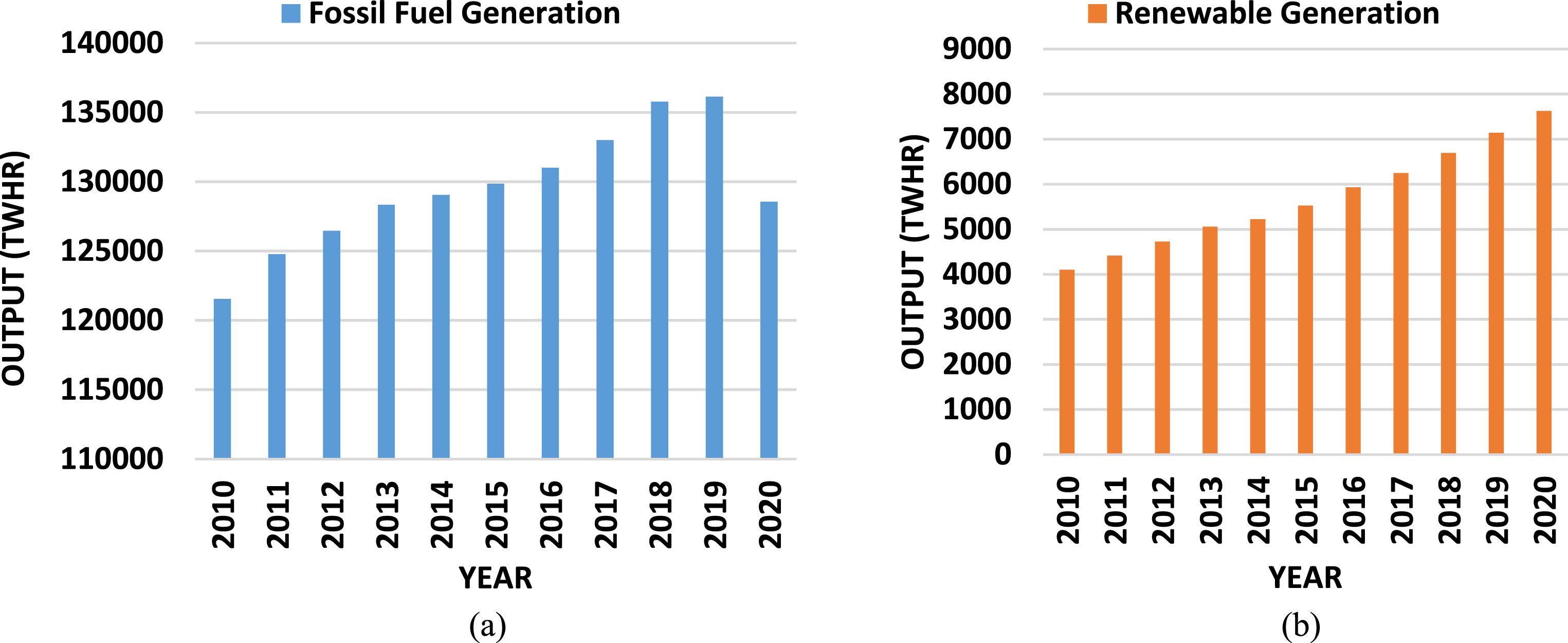}
\caption{An analysis contrasting the production of fossil fuels and renewable energy sources between 2010 and 2020  \cite{ang2022comprehensive}
\label{fig2}}
\end{figure}

The global pandemic of the corona virus made it difficult to obtain current data on the world's electrical generation in 2021 from a number of sources, as \cite{ang2022comprehensive}'s search revealed. The reduction in the production of energy coming from fossil fuels such as and green energy from 2010 and 2020 is seen in Figure \ref{fig2}. The total quantity of fossil fuels utilized to come up with electricity increased considerably, from 121,531 TWh from 2010 to 136,131 TWh starting 2019, before declining in 2020. On the other hand, the quantity of energy produced by green energy has increased significantly, from 4098 TWh by 2010 to 7,140 TWh by 2019. The generation of power from renewable energy (RE) was only 3.26–5.60\% between 2010 and 2020, in comparison to the generation of fossil fuels. 

\unskip

\begin{figure}[H]
\includegraphics[width= 11 cm]{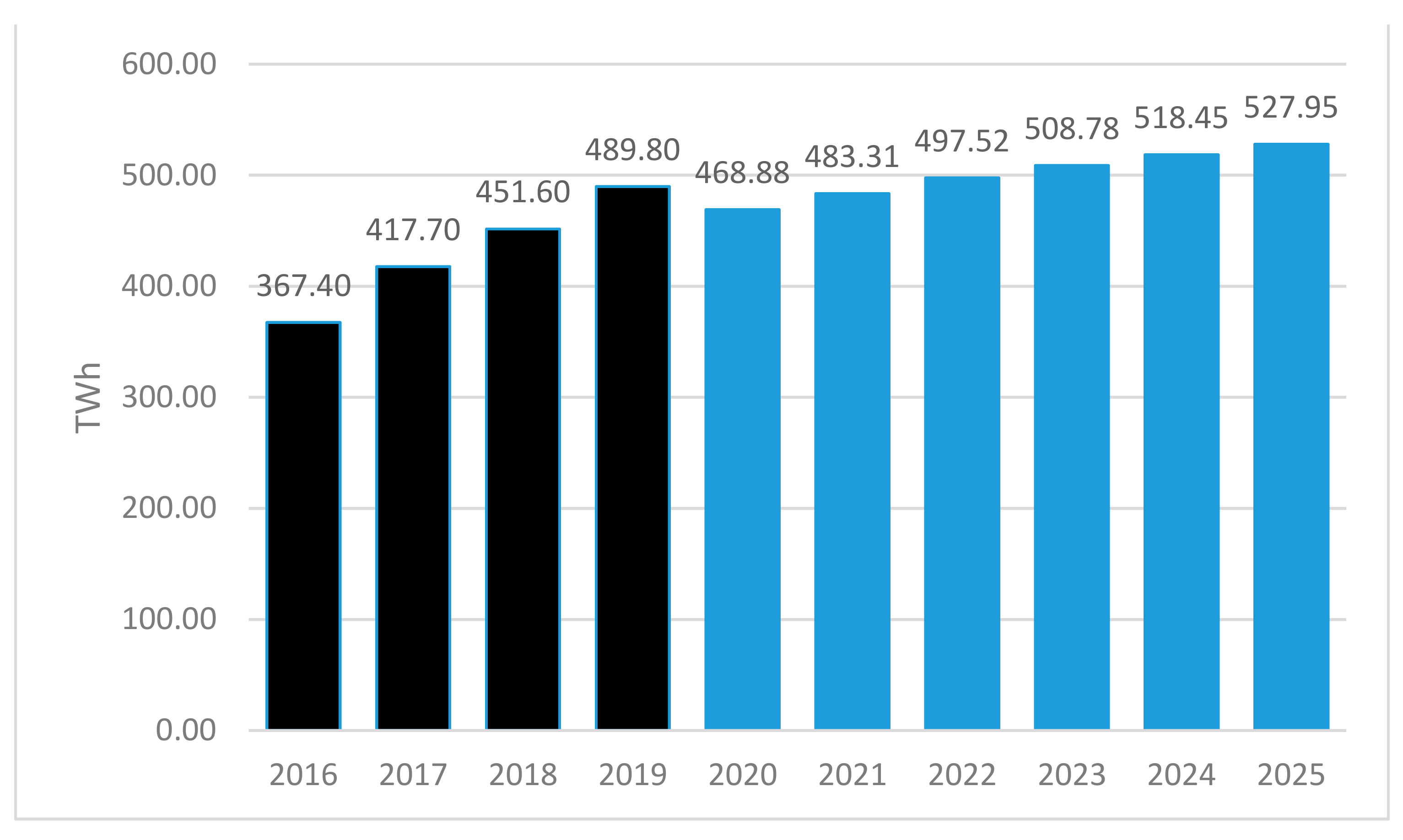}
\centering
\caption{TWh of data on the generation of renewable energy in the United States from 2016 to 2019 and from 2020 to 2025. Authors' calculations serve as the source.  \cite{statistical_review_world_energy_2020}
\label{fig3}}
\end{figure}

Figure \ref{fig3} is showing forcasting that in 2025 the production of Renewable Energy will be reached to 521.95 TWh in USA. This value is incredible with compared to China, Japan, Brazil and Russia's RE production \cite{bhuiyan2021renewable}. 

\unskip

This review paper provides the following contributions in the concerned domain. The contributions can be summarized below:
\begin{enumerate}
    \item \textbf{Comprehensive Review of AI in Renewable Energy (RE):} The paper provides a detailed analysis of how AI is currently applied in the renewable energy sector, offering insights into its use for improving energy efficiency, forecasting, and system optimization.
    
    \item \textbf{Identification of AI-Based Methods:} It highlights nine specific AI-based approaches that support the integration of renewable energy into modern power systems, presenting a clear framework for their application.
    
    \item \textbf{Comparison of AI Techniques and Traditional Methods:} The paper compares the effectiveness of AI techniques with traditional models in areas such as robotics, data processing, and cyberattack prevention, showing how AI provides superior performance.
    
    \item \textbf{Future Directions for AI Integration:} It outlines potential future advancements in the use of AI for renewable energy, focusing on the further optimization of smart grids, energy management, and the development of sustainable power systems.
\end{enumerate}

\subsection{Relevance of Renewable Energy}
Numerous factors that are essential to sustainable development depend heavily on renewable energy \cite{petrovic2020importance}. Primarily, it is essential for lowering greenhouse gas emissions since renewable energy sources, including wind, solar, and hydroelectric power, produce fewer greenhouse gases and offer a sustainable alternative to fossil fuels \cite{perea2021renewable}.Economically speaking, the renewable energy industry stimulates employment, advances technical innovation, and attracts investment, all of which help local economies and the transition to a green economy \cite{petrovic2020importance}. Furthermore, the utilization of renewable energy improves public health by mitigating air and water pollution that arises from burning fossil fuels, consequently relieving respiratory ailments and other health-related issues. Most importantly, it powers socioeconomic growth by enabling fair access to electricity, especially in isolated and underprivileged areas. In addition, the shift to renewable energy drives technological improvement by promoting energy storage, grid integration, and efficiency research and development \cite{alam2020impacts}. 

The major issues can be identified as:

\begin{enumerate}
    \item \textbf{Greenhouse Gas Emissions:} Renewable energy is crucial for reducing greenhouse gas emissions, as sources like wind, solar, and hydroelectric power produce far fewer emissions compared to fossil fuels \cite{petrovic2020importance}.
    
    \item \textbf{Economic Growth:} The renewable energy sector promotes job creation, technological innovation, and investment, contributing to local economies and supporting the transition to a green economy \cite{perea2021renewable}.
    
    \item \textbf{Public Health:} Utilizing renewable energy helps to mitigate air and water pollution caused by burning fossil fuels, which in turn reduces respiratory and other health issues \cite{petrovic2020importance}.
    
    \item \textbf{Socioeconomic Development:} Renewable energy enables equitable access to electricity, particularly in remote and underserved areas, fostering socio-economic growth \cite{alam2020impacts}.
    
    \item \textbf{Technological Advancement:} The transition to renewable energy drives improvements in energy storage, grid integration, and efficiency, spurring further research and development in these areas \cite{alam2020impacts}.
\end{enumerate}

\begin{figure}[H]
\includegraphics[width= 11 cm]{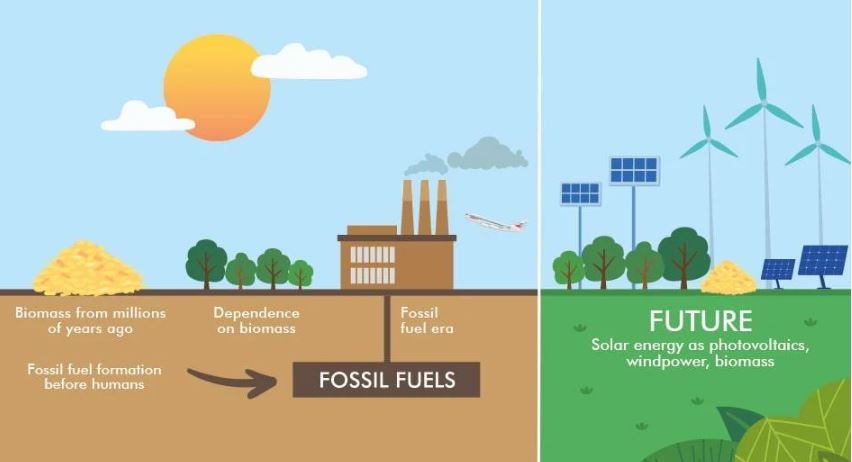}
\centering
\caption{\centering Sustainable eco-friendly solutions for power  \cite{Routledge2021}
\label{fig4}}
\end{figure}  
\unskip

When discussing the issues posed by climate change, many of us immediately consider the impending energy crisis that we will almost certainly confront. Fossil fuel resources are getting harder to come by, but global CO2 emissions are still rising. Future energy requirements and security need the broad use of new renewable energy sources (shown in Figure \ref{fig4}). That still calls for a radical transformation in the way that scientists, students, and world leaders see renewable energy and climate change. Since renewable energy is the way of the future, we must refocus on sustainability via innovation by learning about energy, chemistry, and technology. 

\subsection{Renewable Energy Sector without the influence of Artificial Intelligence}

Renewable energy technologies, including solar, wind, hydropower, geothermal, and biomass, were well-established and continuously developing prior to the incorporation technology artificial intelligence (AI) within the energy industry \cite{onwusinkwue2024artificial}. Developments in these technologies primarily focused on improving efficiency, reducing costs, and addressing environmental concerns. However, despite the advancements brought by AI, there were some limitations and challenges. One of the main lacking aspects was the dependency on traditional methods for data analysis and decision-making, which often lacked real-time insights and predictive capabilities \cite{onwusinkwue2024artificial}. Adoption was hampered by the additional costs associated with infrastructure, data gathering, and algorithm training that came with integrating AI within renewable energy systems, especially in less developed areas. Moreover, there were concerns regarding data privacy, security, and the potential for AI algorithms to exacerbate biases or inequalities in energy access and distribution. Notwithstanding these difficulties, integrating AI has the power to completely transform the renewable energy industry by streamlining processes, boosting productivity, and hastening the shift to sustainable energy sources. Researchers used to concentrate on improving renewable energy sources at that time by creating tools, energy management systems, and appropriate solutions for those that were environmentally detrimental. In this research article \cite{karim2017electricity}, the current state of electricity access is evaluated, problems and obstacles are noted, and the contribution of energy efficiency and renewable energy to the improvement of electricity accessibility in urban poor areas—specifically, the Korail slum of Dhaka—is assessed (2017). This research outcome suggests that rooftop solar PV, electricity generation from solid waste, and replacing incandescent lamps with CFLs are promising approaches to enhance electricity access in the Korail slum, with significant potential for surplus electricity generation and cost savings. A separate study \cite{tao2023environmental} examines the impact of environmental regulations on sustainability, taking into account factors such as research and development in renewable energy (RERD), advancements in technology  sector (TI), and the overall economic performance (GDP). The researchers employed wavelet methods, a specialized mathematical tool, to examine the interconnections between these parameters over various time periods and in diverse manners. They discovered that increasing funding for RERD, implementing more stringent environmental regulations, and promoting the adoption of new eco-friendly technology are crucial for safeguarding the environment. Additionally, they discovered significant correlations between CO2 emissions and factors such as technological advancements and economic expansion.

\subsection{Artificial Intelligence in the Renewable Power Sector}

Digital technology advancements have the potential to drastically change how we use, trade, and supply energy. The new digitalization approach is powered by artificial intelligence (also known as AI) technology \cite{ahmad2021artificial}. The integration of power demand, supply, and sources of renewable energy into the power grid will be autonomously managed by intelligent software that optimizes operations and decision-making. AI will be necessary to accomplish this goal. By linking different gadgets and sensors to the electrical grid, a vast amount of data could be collected. This data, when paired with AI, can give grid operators new insights for more efficient operation control. It allows energy suppliers the flexibility to skillfully adjust supply in reaction in demand.  AI can help with microgrid integration and distributed energy management \cite{Georgiou2019}. Integrating community-scale energy from renewable sources producing units with the main electrical system presents a challenge to maintaining grid balance in energy flow.The AI-powered system of control might potentially significantly alleviate the quality and traffic issues. A trustworthy and long-lasting solution for the renewable energy industry may be provided by the combination of artificial intelligence and intelligent energy storage (IES systems) \cite{Georgiou2019}. 

\begin{figure}[H]
\includegraphics[width= 13 cm]{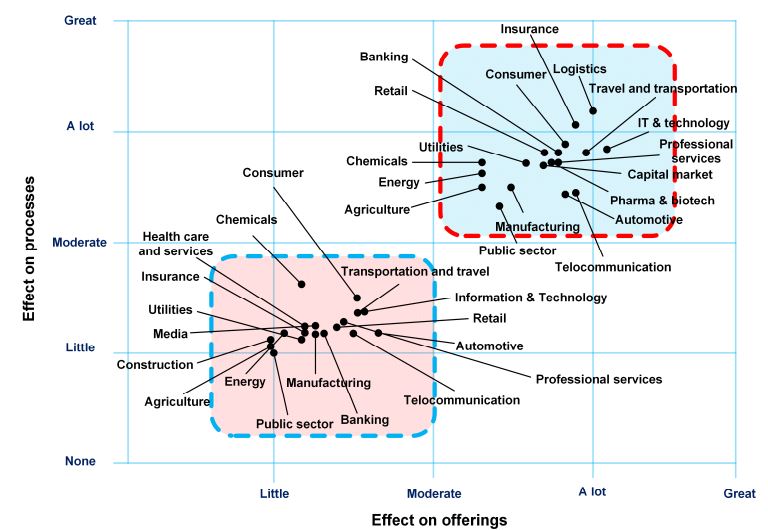}
\caption{\centering Effects of AI in the business along with energy sectors \cite{ahmad2021artificial}
\label{fig5}}
\end{figure}

Figure \ref{fig5} shows how AI is affecting the commercial and energy sectors. It is anticipated that AI will have a greater impact on energy firms than currently anticipated in a number of industries \cite{ahmad2021artificial}. The segment of the red line as shown in Figure \ref{fig5} represents the impact that AI technology on different business types throughout the next five years. The horizontal axis describes the "Effect of offerings," while the vertical axis depicts the "Effect of processes." Although the "Effect of Processes" describes a series of actions taken to achieve a certain goal, the "Effect of Offerings" gives additional opportunities to experience the effect of AI in numerous sectors (agree to or reject as desired). Most businesses expect increased impacts on customer-focused initiatives, manufacturing and operations, supply chain management, including energy information technology (also known as IT). 

\unskip

\begin{center}
  \begin{figure}[H]
    \includegraphics[width=10cm]{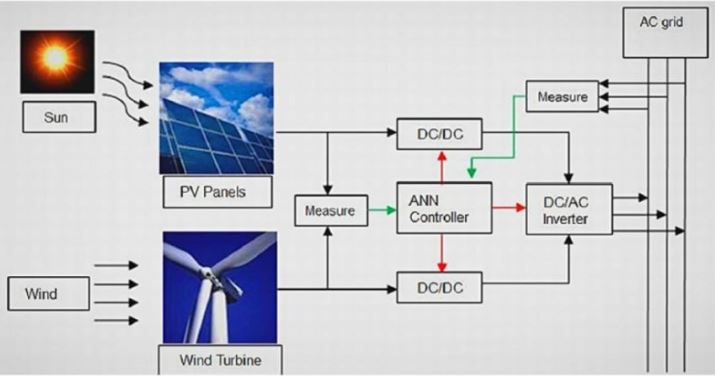}
    \centering
    \caption{\centering AI and ML-based PV-wind system control \cite{Lahmadi2023}}
    \label{fig6}
  \end{figure}
\end{center}

Control solutions that make use of data-driven models of the wind-photovoltaic system employ AI and ML algorithms (Shown in Figure \ref{fig6}). A wind turbine's power production may be predicted by an AI-based control algorithm based on meteorological factors like wind direction and speed \cite{Lahmadi2023}. Similar to this, a control system based on machine learning may be created to forecast a photovoltaic panel's power production depending on variables like temperature and sun irradiation. The wind-PV system's performance may then be optimized using these forecasts.
\unskip
Data-driven models such as support vector machines (known as SVM), random forest (known as RF), k-nearest neighbors (known as SVM k-NNs), regression trees, multiple linear regression, and gradient boosting (known as GB)  are commonly used in the field of renewable energy (RE) prediction, leveraging machine learning (ML) techniques. Deep learning models such as long short-term memory (also known as LSTM), gated recurrent units (also known as GRUs), and deep neural networks (also known as DNNs) have been used to predict consumption of power  and renewable energy generation across different time periods, yielding outstanding results \cite{khan2021db}, \cite{nam2020deep}. In addition, the combination of LSTM with autoencoders (AE) has produced positive outcomes \cite{gensler2016deep}.

\subsection{Survey Framework of this Research }

The employment of AI techniques in the energy sector is the main topic of this study. This study aims to establish a realistic baseline so that researchers and readers may assess each other's AI initiatives, ambitions, novel, cutting-edge applications, challenges, and future trajectories. Three main topics were covered: (i) utilising current developments in AI technology to create renewable power; (ii) use AI to forecast renewable energy; and (iii) optimizing energy systems. This research also looked at how AI techniques outperform traditional models in the areas of controllability, huge data handling, cyberattack protection, smart grid, Internet of Things, robotics, conservation of energy optimization, automated maintenance control, as well as computing efficiency. Another important area of this study is the role AI will play in the emerging energy sector.

\begin{figure}[H]
\includegraphics[width= 14 cm]{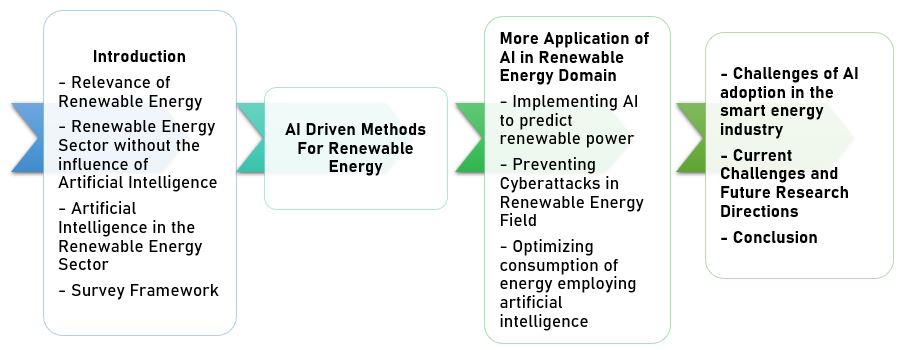}
\caption{\centering Survey framework of proposed survey research. 
\label{fig6.1}}
\end{figure}  
\unskip

The survey structure of this research is organized in a strategic and logical way to cover all the important issues which is shown in the Figure \ref{fig6.1}. This is how the remaining portion of the paper is structured. 

\section{AI Driven Methods For Renewable Energy} \label{sec3}
AI-driven methods for renewable energy are transforming the way power systems operate by enhancing efficiency and reliability. Machine learning algorithms, such as neural networks and reinforcement learning, are widely used to forecast energy generation, optimize grid management, and improve demand-response systems. These techniques help in accurately predicting renewable energy output based on weather conditions, ensuring better integration of solar and wind power into the grid. AI also plays a key role in energy storage management, optimizing the use of batteries and other energy storage systems. Additionally, AI-driven models contribute to reducing energy waste and improving the overall sustainability of renewable energy systems.

\subsection{Forecasting Renewable Energy Generation using Deep Learning}

\begin{figure}[H]
\includegraphics[width= 11 cm]{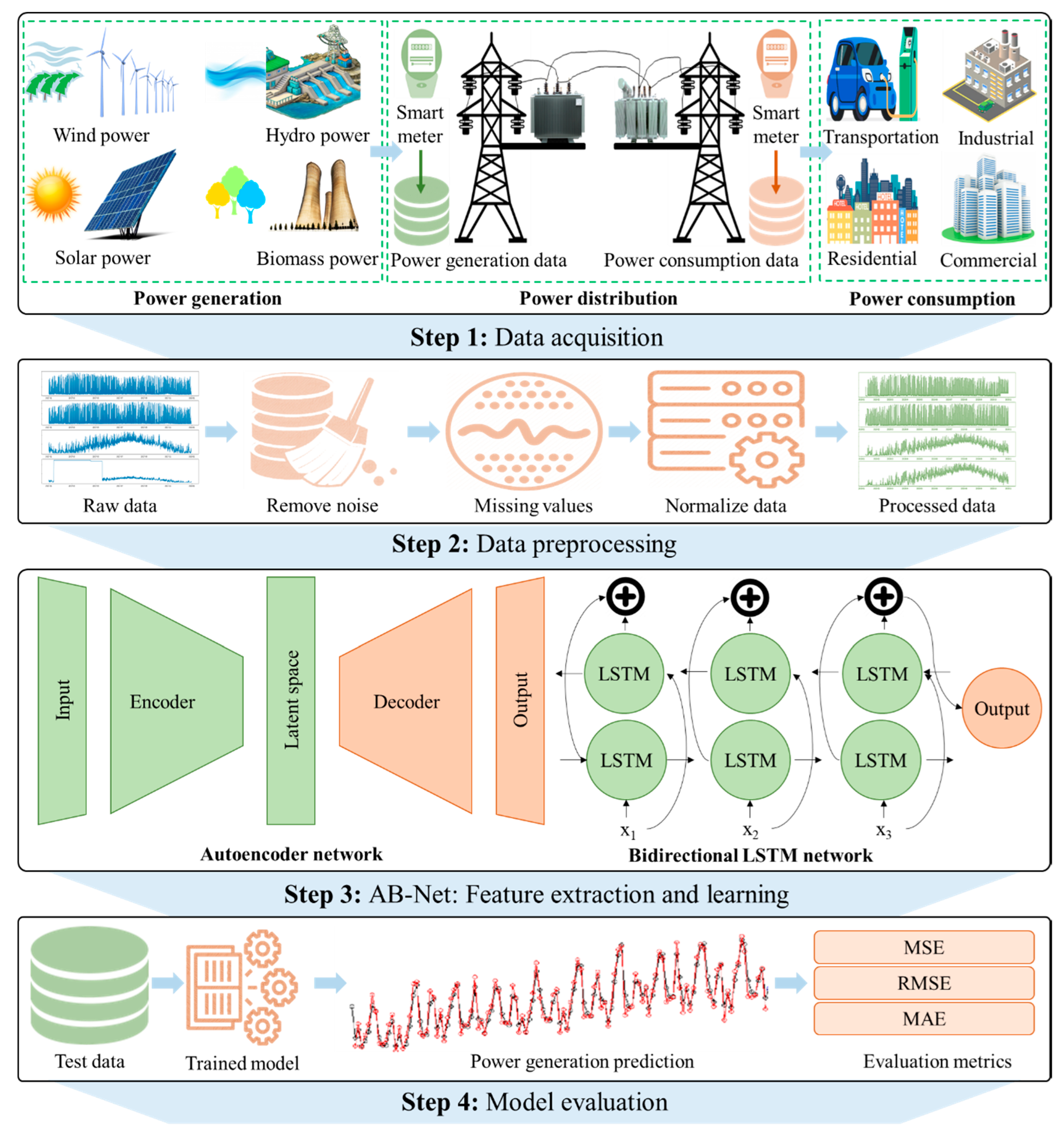}
\centering
\caption{Design presented by authors adheres to a well-defined framework. Data on electricity generation is acquired during Step 1. The second step entails preparing this data. During  third step, features are retrieved as well as then transmitted through BiLSTM in support of purpose of decoding. At the final step, Step 4, model that has been trained is used to generate predictions. These predictions are then assessed using simple error measures and shown visually through graphs. \cite{khan2021ab}
\label{fig7}}
\end{figure}

The authors \cite{khan2021ab} suggest the 'AB-Net' architecture allowing one-step modeling of renewable energy (RE) generation, which combines autoencoder (AE) with bidirectional long-short-term (BiLSTM). By acquiring data from various RE sources and deep preprocessing, the model extracts features and forecasts RE generation.The wind database and the solar dataset, two publicly available datasets, were used to test and evaluate the effectiveness of the proposed strategy. The model consists of three main steps: data acquisition from various renewable energy sources, deep preprocessing to clean and normalize the acquired data, and feature extraction and forecasting using AE and BiLSTM which is shown in Figure \ref{fig7}. 

\begin{equation}
\begin{aligned}
h_t &= \sigma_h(W_{xh} X_t + W_{hh} h_{t-1} + b_h)
\end{aligned}
\end{equation}

\begin{equation}
\begin{aligned}
y_t &= \sigma_y(W_{hy} h_t + b_y)
\end{aligned}
\end{equation}

Equation (1) describes the state calculation of a hidden layer within LSTM network. The hidden state \(h_t \) is kept up-to-date and at every interval \(t \) depending on the input \(x_t \) that is currently being fed into the layer as well as the prior hidden state \(h_{t-1} \). The activation function is represented by \(\sigma_h \) in Equation (1), the weight matrix connecting the input to the hidden layer is represented by \(W_{xh} \), the weight matrix with regard to recurrent connections of hidden states is represented by \(W_{hh} \), and the bias vector for the hidden layer is represented by \(b_h \).

\unskip

When compared to other models such an encoder-decoder (ED), CNN-BiLSTM, and BiLSTM, AB-Net performed better with regard of mean absolute error (in short MAE), mean squared error (in short MSE), as well as root mean square error (in short RMSE). The solar dataset's mean squared error (MSE), mean absolute error (MAE), and root mean square error (RMSE) were 0.0106, 0.1028, along with 0.0743, in that order. The comparable values for the wind dataset were 0.0004, 0.0189, along with 0.0109, in contrast. This hybrid network enhances energy management efficiency by facilitating enhanced integration, trading, and control of renewable energy installations. 

\subsection {Big Data and Machine Learning for Smart Grid}
\begin{figure}[H]
\includegraphics[width= 11 cm]{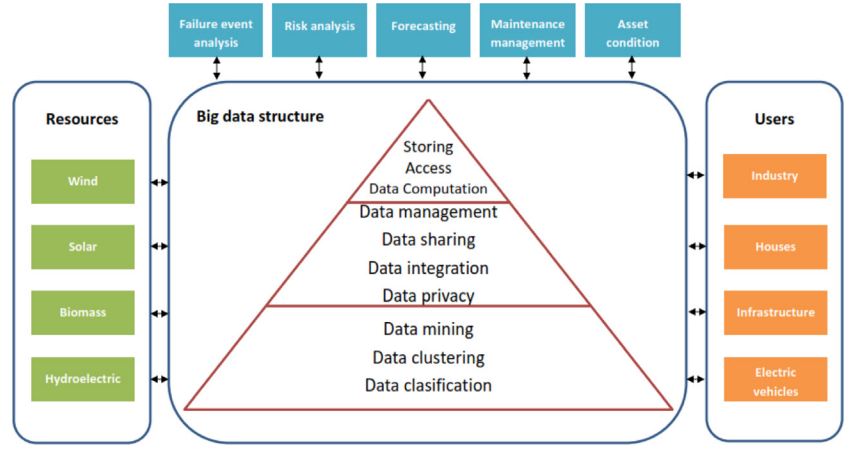}
\centering
\caption{\centering A big data structure for a power utility using renewable energy \cite{mostafa2022renewable}.
\label{fig8}}
\end{figure}

Three layers compose the big data framework, as seen in Figure \ref{fig8}. The purpose of the higher, simpler layer is to store, retrieve, and perform calculations on data. Data management, sharing, and integration across many applications and domains are the responsibilities of the intermediate layer; data privacy is a major concern in this layer. The data mining platform uses data fusion technologies to preprocess data at the lowest and deepest layer.

\unskip

With an emphasis on system stability prediction, the research study \cite{mostafa2022renewable} investigates the implementation of big data analytics (BDA) between intelligent grids and energy from renewable sources power plants. A framework is developed and applied to a dataset of 60,000 instances using various machine learning methods, achieving high accuracy in predicting stability. Three ML models were used: deep learning with a typical neural network, random forest classifier, as well as decision tree. The goal of the first model, which is connected to classification model, was to determine whether or not the smart grid system will be regarded as stable. The fourth ML model estimates the equilibrium differential equation root using a penalized linear regression. Notably, convolutional neural networks (CNN) obtain 87\% accuracy for classification, compared to penalized linear regression model's 96\% accuracy. The main contribution lies in demonstrating the effectiveness of BDA in enhancing decision-making and system stability in smart grid environments, paving the way for future research on larger datasets and broader applications.

\subsection{Using Machine Learning Models For Local Wind Power Forecasting (Neural Network Architecture and Tree-Based Approaches)}

\begin{figure}[H]
\includegraphics[width= 11 cm]{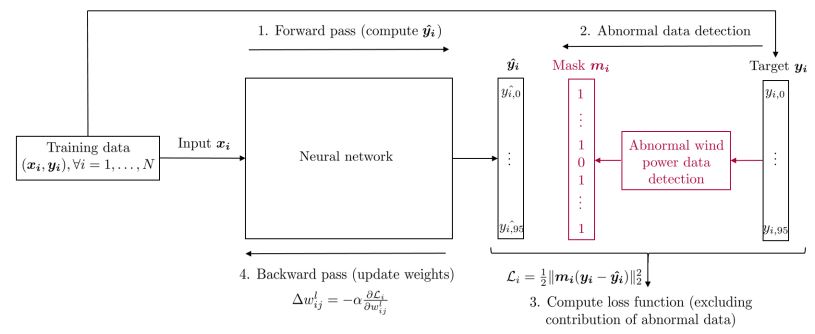}
\caption{Method to train neural networks when anomalous data from wind power is present. \cite{greve2020machine}.
\label{fig9}}
\end{figure}

Local prosumers and consumers congregate in communities where they can collaborate or engage in competition to achieve shared goals such as reducing greenhouse gas emissions and/or power prices. Authors of the research, \cite{greve2020machine}, propose data analytics modules for Renewable Energy Communities (RECs) to optimize resource usage and minimize electricity bills. It develops a day-ahead wind power forecasting algorithm using Machine Learning techniques, improving accuracy by 10\% and offering representative consumption profiles for community members. The overall process is shown in Figure \ref{fig9}, where abnormal wind power data is recognized in the target vector and a "forward pass" is performed to compute the actual forecast. The neural network's parameters are changed in the backward pass after the loss function has been calculated in the third and final step. 
\unskip
The research introduces innovative approaches for wind power prediction, anomaly detection in data, and the generation of consumption profiles, facilitating the seamless integration of renewable energy into community grids. Implemented initially in a pilot Renewable Energy Community (REC) in Belgium, these modules aim to synchronize energy consumption with local generation, thereby improving coordination and leveraging favorable pricing mechanisms. Based on data from the E-Cloud pilot project, the ENSEMBLE model has exhibited superior forecasting accuracy. It aggregates predictions from four distinct machine learning models: random forests, gradient boosting decision trees, multilayer perceptrons, and bi-directional LSTMs, providing robust forecasting outcomes. The training procedure has been modified to account for the automated detection of anomalous wind power data samples, which has significantly increased forecasting accuracy. a method that uses dynamic time warping to create community members' representative power usage profiles. 
\begin{figure}[H]
\includegraphics[width= 11 cm]{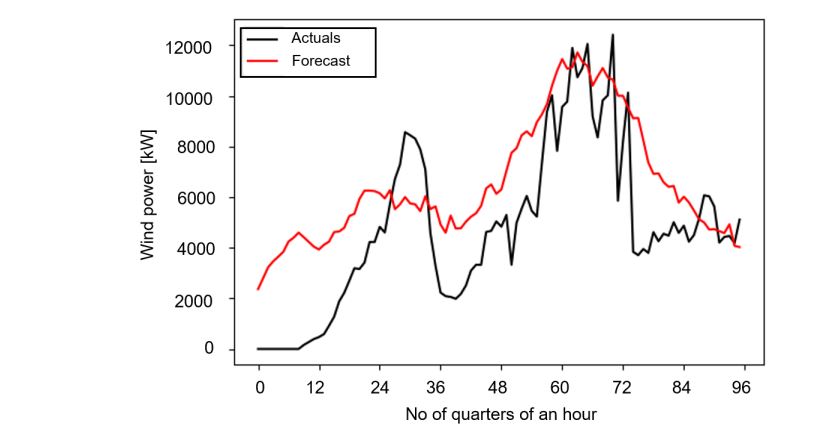}
\caption{Time series reveals the test set's actual and predicted wind power for a single day.
\cite{greve2020machine}.
\label{fig10}}
\end{figure}  
\unskip
In Figure \ref{fig10}, the graph shows the actual wind power generation (in black) alongside the wind power forecast (generated using the ENSEMBLE model) over time, depicting data from a random day within the test set.

\subsection{Using Machine Learning and Ensemble Feature Selection to Forecast Solar Radiation}

\begin{figure}[H]
\includegraphics[width= 6 cm]{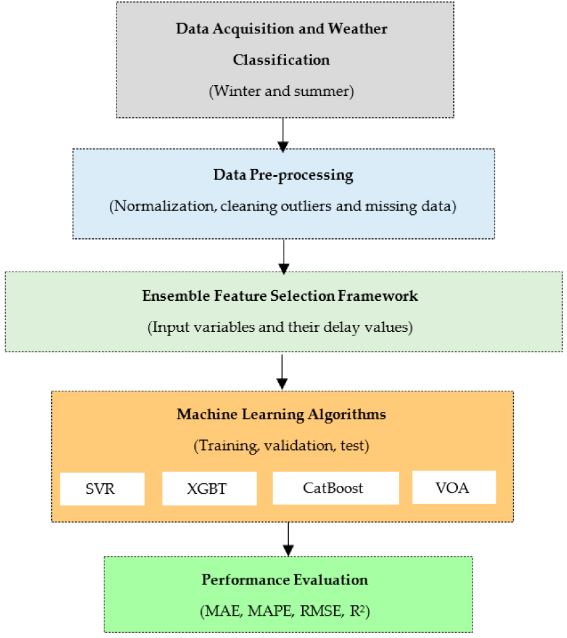}
\centering
\caption{\centering Diagram illustrating the suggested methods
\cite{solano2022solar}.
\label{fig11}}
\end{figure}

Precise prediction of  radiation from solar is vital for the reliable operation of power systems that rely heavily on photovoltaic generation. This study \cite{solano2022solar} assesses the performance of machine learning (ML) methods in predicting solar radiation. The algorithms evaluated include Support Vector Regression (in short SVR), Categorical Boosting (in short  CatBoost), Voting-Average (in short  VOA) , and Extreme Gradient Boosting (known as XGBT). A novel approach is introduced to choose significant variables and their previous observations, hence improving the accuracy of predicting. The most related internal and extrinsic inputs, together with their historical observation values, are selected using an ensemble feature selection technique that incorporates Pearson's coefficient, mutual information, RF, and alleviation. The suggested method for estimating solar radiation consists of five important phases, as illustrated in Figure \ref{fig11}. Tested on real data from Salvador, Brazil, VOA outperforms other algorithms in all prediction time horizons. Contributions include comparing ML algorithms, introducing CatBoost for solar forecasting, and proposing an ensemble feature selection method. Results show improved accuracy and highlight the importance of selecting appropriate input variables and past observations for solar radiation forecasting. For the summer and winter datasets, VOA had the faster training speed (183.93 s and 14.76 s, respectively). Being more sophisticated, VOA combines SVR, XGBT, and CatBoost. Further research is recommended to optimize parameter selection and explore applications in different locations and forecasting domains.
\unskip

\subsection{Time Delay Neural Network (known as TDNN): The Best Hybrid AI Approach for Smart Grid Management of Energy}

In their study \cite{conte2022new}, the authors propose an optimal hybrid AI strategy to enhance energy management efficiency in smart grids like Renewable Energy Communities. This approach employs a Time Delay Neural Network to predict future energy characteristics within the community. Subsequently, a stochastic model predictive control utilizes these predictions to optimize community operations through effective management of battery energy storage systems. Results from forecasting using a publicly available dataset show Mean Absolute Errors of 1.60 kW for solar generation, 2.15 kW for total energy consumption, and 0.30 kW for common services over a 24-hour prediction horizon. Comparative analysis reveals that the predictive management model significantly increases revenue compared to competitors. Specifically, leveraging forecasts improves total income by 18.72\% compared to employing the same management system without utilizing forecasting.

\subsection{Techniques for Detecting Transient Stability Using CNN-LSTM}
\begin{figure}[H]
\includegraphics[width= 10 cm]{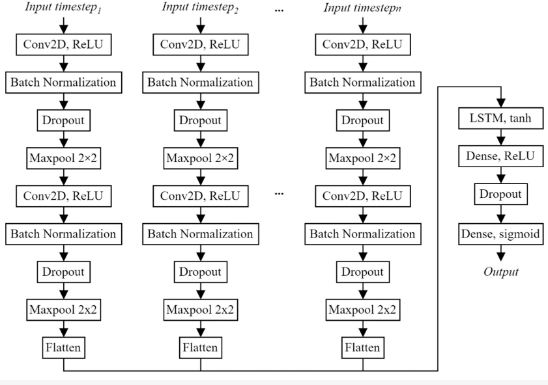}
\centering
\caption{\centering CNN-LSTM design that is being proposed \cite{azhar2022development}.
\label{fig12}}
\end{figure}

When a disturbance arises, electric power systems are more vulnerable to instability because they function closer to their limitations. Problems with transient stability are particularly common. \cite{azhar2022development}, the research introduces a novel CNN-LSTM model (shown in Figure \ref{fig12}) to detect transient stability in electric power systems using historical data events. The model addresses challenges in power system operation, taking into account factors such as noise, delays, data measurement losses, line outages, and scenarios involving the integration of variable renewable energy (VRE). Achieving over 99\% accuracy in identifying stable and unstable conditions, the proposed model outperforms conventional CNN and LSTM approaches in terms of computation time, averaging 190.4 seconds. By training on synthetic data and considering various system parameters, this model demonstrates promise for enhancing transient stability detection in real-world power systems, albeit with challenges in implementation and optimization. In terms of computation time and performance, the CNN-LSTM approach outperforms the CNN and convLSTM approaches.
\unskip

\section{Additional Applications of AI in the Renewable Energy Sector}

The energy industry's significant development in the utilization of AI, Internet of Things, and smart grid infrastructure control has led to a study of cyber-attacks and a rapid rise in the number of solutions \cite{sani2019cyber}. When it comes to load balancing renewable energy, automation is particularly crucial since it ensures grid stability by matching distributed generation to load \cite{vernotte2018load}. The actions of rogue or rouge actors combined with inadequate system design, implementation, or configuration pose a threat to load balancing and the smart grid as a whole. Before a cybercatastrophy occurs, it is really imperative to perform thorough investigation about viability and urgency of these possible dangers \cite{vernotte2018load}.

\subsection{Implementing AI to predict renewable power}

Researchers and the energy business have paid close attention to big data as well as artificial intelligence (AI) in recent years. This interest stems from two key factors: (i) growing processing power and (ii) the exponential global data expansion \cite{mazzeo2021artificial}.The authors of the study \cite{mazzeo2021artificial} recommend employing Artificial Neural Networks (ANNs) to model and size a clean energy community (CEC) incorporating energy storage devices, electric vehicle charging stations, and a PV-wind hybrid system to meet district energy demands. One ANN is dedicated to estimating the grid energy indication factor, while another predicts energy performance indicators such as the fulfilled load fraction and utilization factor of generated energy. ANNs are trained extensively across diverse weather conditions, variable power system configurations, and a range of electrical demands, using advanced models for each CEC component to directly forecast annual energy performance without dynamic system simulations. The Garson algorithm assesses the impact of each input on the outputs. Optimized ANNs, characterized by a single hidden layer with twenty neurons, demonstrate highly accurate predictions for CECs compared to those used during training.

\subsection{Preventing Cyberattacks in Renewable Energy Field}
The clever renewable energy to cyberattacks, opening a new avenue for hackers to exploit the weaknesses in renewable energy \cite{rekeraho2024cybersecurity}. This clearly impacts smart renewable energy's operational and economical aspects \cite{rekeraho2024cybersecurity}. Recent advancements in communication networks, monitoring, smart control systems, and the widespread adoption of Internet-based architectures have led to significant transformations in Cyber-Physical Power Systems (CPPSs), despite encountering some conflicting perspectives \cite{ghiasi2023comprehensive}. Cyber components and power portions are typically connected in these architectures. CPPSs deal with recently developed problems like as security, vulnerability, resilience, stability, and dependability. Assessing, evaluating, and offering remedies to lessen or resolve these issues heavily rely on precise modeling techniques and investigating the interplay processes related to Smart Grids' (SGs') cyber-security. Many studies focused on creating reliable methods for identifying and/or defending against cyberattacks. These included studies on detection mechanisms that use federated learning techniques, perfect or imperfect false data attacks, false data-driven or data-injection attacks, and designing resilient communication networks  \cite{ghiasi2023comprehensive}. 

The research paper \cite{ahmadi2022new} presents an innovative data-driven method for detecting false data injection (FDI) attacks in energy forecasting systems. This approach is critical to ensuring the reliability of power systems, particularly as reliance on renewable energy and information technology continues to grow. Utilizing cross-validation, least-squares, and z-score metrics, the proposed mechanism effectively identifies intrusions without relying on system models or parameters, enhancing system resilience. Through experiments with tree-based wind power forecasting models, the approach successfully detects and removes corrupted data, restoring forecasting accuracy and generalizability.

\begin{figure}[H]
\includegraphics[width= 10 cm]{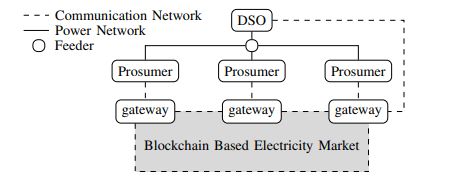}
\centering
\caption{\centering Infrastructure for the decentralized transactive energy system \cite{barreto2020cyber}.
\label{fig13}}
\end{figure}

The rising prevalence of energy storage devices and the expanded utilization of intermittent renewable energy sources are driving substantial transformations in power systems \cite{barreto2020cyber}. These developments prompt smart-grid operators to imagine a day when microgrids would facilitate peer-to-peer energy trading, which will result in the creation of transactive energy systems. Due in major part to their remarkable durability, blockchains have attracted a lot of interest from academics and business for their possible use in decentralized TES. The research \cite{barreto2020cyber} explores blockchain-based Transactive Energy Systems (TES) and introduces a new type of attack aimed at gateways between prosumers and the system. Figure \ref{fig13} illustrates the general architecture of the decentralized TES. The study develops a detailed model of blockchain-based TES, examines various threat scenarios, and proposes strategies for mitigating these attacks. Experimental results using GridLAB-D and a private Ethereum network show that even basic attacks can disrupt market equilibria. However, mitigation measures such as detection and gateway switching are effective in countering these threats, thereby maintaining the system's integrity and resilience.

\unskip

\subsection{Optimizing consumption of energy employing artificial intelligence}

The subject matter "Cost Savings" via AI-powered energy efficiency in smart systems is one of the advantages. Lower operating expenses and electricity bills are directly correlated with decreased energy use. A variety of methods are included in AI technology, such as predictive modeling, data analytics, and machine learning. AI-powered systems may learn from data trends, make wise judgments, and optimize energy use in real time thanks to their capabilities. When it comes to optimizing energy usage in smart buildings, AI has several advantages. It makes it possible to gather and analyze enormous volumes of data from different sensors, meters, and Internet of things devices. HVAC (heating, ventilation, and air conditioning) systems are a primary contributor to the energy consumption of buildings. AI-based algorithms can optimize HVAC operations by considering factors such as occupancy, weather conditions, and thermal comfort requirements.

In this paper \cite{ilbeigi2020prediction}, a robust ANN network trained with data from an Iranian research center is combined with EnergyPlus software to suggest a dependable technique for optimizing energy usage in buildings. Genetic algorithms are used to optimize a building's energy usage. Highly efficient features are evaluated in the building's energy usage. Through the use of sensitivity analysis, the elements that have the greatest impact on energy consumption are determined; the most important ones are the wall U-value and the number of occupants. The proposed ANN model accurately predicts energy consumption, enabling optimization through a Genetic Algorithm-based approach, resulting in a substantial 35\% reduction in energy usage. Overall, the study provides valuable insights into optimizing energy consumption in buildings, offering a promising methodology for similar applications.

\section{Obstacles to Implementing AI in the Smart Energy Sector}

The smart energy sector encounters numerous obstacles in adopting AI, such as insufficient or poor-quality data, challenges with AI network parameters, technical infrastructure issues, a lack of skilled professionals, integration difficulties, as well as risks and legal compliance concerns. Additionally, failure detection and diagnosis in energy systems pose further complex problems \cite{guo2019expert}. Several studies highlight that insufficient information and data security are significant challenges for energy systems \cite{zhao2013intelligent}. The performance and reliability of energy systems are adversely affected by inadequate sensors, controllers, and controlled devices used for data estimation and system operation. The energy market also faces complications due to the intricate relationships and strong interdependencies within the power grid, as well as the high dimensionality and complexity of large-scale simulation grid data \cite{tang2018framework}. Furthermore, integrating renewable energy sources such as solar and wind with AI in grid operations presents additional complexities \cite{puri2019hybrid}.

The integration of artificial intelligence (AI) in the smart energy sector encounters numerous obstacles. One significant issue is the lack of essential AI expertise among decision-makers, which impedes their comprehension of AI's potential advantages \cite{ahmad2021artificial}. Moreover, there is an evident gap in practical skills necessary for creating dependable AI applications that deliver tangible real-world benefits, causing power companies to hesitate in embracing new approaches. Efforts to modernize are also hampered by outdated power system infrastructure \cite{ahmad2021artificial}, with utility companies facing difficulties in managing extensive amounts of dispersed and unstructured data. Economic pressures also impede the integration of innovative energy technologies due to high costs and resource requirements. Furthermore, decentralization and diversification of energy supply, coupled with emerging AI technologies, present complex challenges for energy production and distribution worldwide. Dependency on cellular technologies limits AI's potential \cite{ahmad2021artificial}, especially in developing economies. Furthermore, AI-based applications frequently function as opaque systems for consumers, leading to worries about transparency and security, especially given the rising cyber threats within the digital energy infrastructure. Overcoming these challenges is crucial to fully harness AI's potential in transforming the smart energy sector.

\section{Current Challenges and Future Research Directions}\label{sec4}

The incorporation of AI into research on renewable energy presents a number of obstacles as well as opportunities for further investigation. First and foremost, attention needs to be paid to the complexity and expense of AI algorithms employed in the design, optimization, and administration of renewable energy systems. Simplifying and making these models cost-effective is crucial for widespread adoption and scalability.

Critical challenges in real-time include automating circuit breakers and relays under fault conditions, detecting and preventing the impact of lightning on transmission lines, improving the efficiency of geothermal plants using IoT and AI, and regulating demand for load for wind and solar power farms under extreme weather.

Regulators and policymakers must recognize AI's systemic impact on reducing greenhouse gas emissions and ensure flexible policies to facilitate the rapid deployment of AI in the domain of energy. Incorporating ethical implications of AI in corporate social responsibility standards is essential for responsible AI deployment. Establishing participatory mechanisms to understand stakeholder expectations and limitations in AI application cases is imperative. It will be extremely beneficial to develop AI applications that precisely detect energy cost and effectiveness reductions for both individuals and enterprises.

Future research should focus on improving system data accessibility and openness, safeguarding models based on artificial from security threats, and optimizing AI's integration with additional technologies like robots, sensors, and the Internet of things. AI may play a significant role in the grid's integrating sources of renewable energy and promotion of operational independence with distributed sources of energy and microgrids. AI may be introduced to developing power markets with the help of international financial institutions, enabling a sustainable energy transition.

Last but not least, realizing AI's promise in energy systems calls for a multifaceted strategy, assessing the viability of integrating AI, finding interesting AI methodologies, and offering practical insights for improving performance while pleasing a range of multidisciplinary audiences. In order to solve the difficulties posed by global climate change and achieve sustainable energy transitions, it is imperative that comprehensive frameworks and techniques for the efficient integration of AI in energy systems be developed.

From revolutionizing medical diagnoses \cite{10054934}, \cite{biswas2023hybrid}, \cite{biswas2022mri}, \cite{biswas2023active}, \cite{biswas2023generative} to enhancing personalized learning in education, and optimizing transportation networks for efficiency, artificial intelligence has been instrumental across various sectors \cite{irbaz2022real}, \cite{biswas2021efficient}. Now, with recent advancements, the power sector is harnessing AI's potential to improve grid management, forecast energy demands, and enhance overall operational efficiency. Researchers believe that combining AI with renewable energy sources will lead to significant improvements in sustainability and energy efficiency. AI has promise for improving forecasting precision, grid stabilization, renewable energy system optimization, and the provision of more intelligent managing energy choices. Given AI's capacity for massive data analysis and real-time decision-making, we see a future in which renewable energy sources will predominate in supplying the world's energy needs while minimizing its negative environmental effects. The combination of AI with renewable energy is expected to provide a more resilient and sustainable energy environment, paving the way for a more environmentally friendly future.

\begin{longtable}{ @{} p{1.2cm}  p{0.7cm}  p{4.5cm}  p{3cm}  p{2.5cm}  @{}}
\caption{An overview of this survey article that includes the methodology and results of previous studies}\label{tab1} \\
\toprule
Reference & Year & Methodology Model & Data Source & Research Outcome \\
\midrule
\endfirsthead

\multicolumn{5}{c}%
{{\tablename\ \thetable{} -- continued from previous page}} \\
\toprule
Reference & Year & Methodology Model & Data Source & Research Outcome \\
\midrule
\endhead

\midrule
\multicolumn{5}{r}{{Continued on next page}} \\
\endfoot

\bottomrule
\endlastfoot

\textbf{\textit{khan et al. }}  & 2021   & Used DL method AB-Net for forecasting of renewable energy generation & Solar \cite{datagokr}, Wind \cite{nrelwind} &  MSE 0.0004, RMSE 0.0189. 
\\

\textbf{\textit{mostofa et al. }}   & 2022   & Used DT, RF, CNN, and linear regression for smart grid management & Smart Grid \cite{8587498} &  Linear regression 96\% accuracy. \\

\textbf{\textit{greve et al. }}  & 2021  & Day-ahead wind power prediction using ML techniques (RF, GB, MLP, BiLSTMs) & Meteo Data \cite{shen2018combined}, \cite{greve2020machine} &  RMSE 2327 kW. \\

\textbf{\textit{solano et al. }}  & 2023  & XGBT, SVR, VOA, and CatBoost for solar radiation forecasting & Brazilian INMET \cite{inmet} & Winter-VOA MAE 0.2417, Summer-VOA MAE 0.2877. \\

\textbf{\textit{conte et al. }}   & 2022  & Time Delay Neural Network for energy management  & PV generation plant data \cite{ramos_energy_2021} & Income enhanced by 18.72\%. \\

\textbf{\textit{azhar et al. }}  & 2022  & CNN-LSTM model to detect transient stability in power systems  & DigSILENT synthetic dataset \cite{izzuddinfa_transient_stability} &  Stability detection >99\%. \\ 

\textbf{\textit{mazzeo et al. }}  & 2022  & ANN for EV charging station and HRES performance assessment & Climate data from 48 locales \cite{mazzeo2021artificial} &  R2 values: SLF 0.952, GEIF 0.925, UF 0.942.\\ 

\textbf{\textit{barreto et al. }}  & 2021  & TES based on blockchain, exploring cyberattacks & Prosumers' historical bid data &  Threats reduced via gateway switching.\\ 

\textbf{\textit{ilbeigi et al. }}  & 2020  & Genetic algorithms to optimize building energy usage & Weather data loaded into Metronome &  ANN model 99.9\% accuracy in energy use remodeling.\\ 

\end{longtable}

\section{Conclusion} \label{sec5}

Our study paper provides a thorough analysis of the application of artificial intelligence (AI) methods utilized renewable energy (in short RE) systems. Nine AI-based techniques for modern power systems—which address topics including renewable power production, energy forecasting, and system optimization—are found by means of a comprehensive literature study. The reviewed studies show how various AI techniques, like as deep learning, random forests, decision trees, and neural networks, are effective in addressing a variety of RE management issues.

The surveyed research demonstrates the potential of AI in enhancing controllability, data handling, cyberattack prevention, and smart grid implementation within the energy industry. Furthermore, the superiority of AI models over conventional approaches is highlighted, particularly in terms of accuracy, efficiency, and adaptability to complex scenarios. The paper describes the potential for integrating AI into energy through renewable sources systems in the future and emphasizes the need for further research and development in this rapidly evolving field. Subsequent investigations ought to concentrate on enhancing AI algorithms, merging AI with other state-of-the-art technologies such as blockchain and IoT, and devising legislative structures to promote the application of AI in the field of energy.

Overall, this survey paper contributes valuable insights into the futuristic AI applications in renewable energy as well as provides roadmap for future research endeavors aimed at leveraging AI for sustainable energy transition and environmental conservation.

\section*{Declarations}

\subsection* {Funding}A portion of the funding for this study comes from NSF Grant No. 2306109 and DOEd Grant P116Z220008 (1). All opinions, findings, and conclusions are presented by the author(s); the the sponsoring organization may not always agree with them.

\subsection*{Conflict of interest} The authors have no conflicts of interest to declare.

\subsection*{Ethics approval} There isn't any original study involving humans, animals, or sensitive data in this review publication.

\subsection*{Consent to participate} The reviews and data used in this analysis were from previously published, accessible to the public literature. 

\subsection*{Consent for publication} The final piece of work has been reviewed and approved for obedience to ML journal by every author included in this review research. Each author has contributed significantly to the manuscript's composition, research, and modification. Each author has carefully examined and approved the material in this analysis, and they have all expressly consented to its publication.

\subsection* {Availability of data and materials} The study's references provide the information and resources that were used.

\subsection* {Code availability} The review process involves a comprehensive analysis of publicly available literature, with due acknowledgment given to the original authors of any techniques or algorithms included. We have endeavored to provide clear references and citations so as to aid readers in identifying the original works for closer examination.

\subsection* {Authors' contributions}
 Conceptualization: P.B., A.B., M.A.A.N.; Methodology: A.R., A.B., M.A.A.N; Formal analysis and investigation: A.B., M.A.A.N; Writing - original draft preparation: A.B., M.A.A.N., A.R., P.B.; Writing - review and editing: M.A.A.N., A.R.; Funding acquisition: K.D.G., R.G.; Resources: A.R., P.B.; Supervision: K.D.G., R.G. All authors have read and agreed to the published version of the manuscript.



\bibliography{sn-bibliography}


\begin{thebibliography}{61}
\ifx \bisbn   \undefined \def \bisbn  #1{ISBN #1}\fi
\ifx \binits  \undefined \def \binits#1{#1}\fi
\ifx \bauthor  \undefined \def \bauthor#1{#1}\fi
\ifx \batitle  \undefined \def \batitle#1{#1}\fi
\ifx \bjtitle  \undefined \def \bjtitle#1{#1}\fi
\ifx \bvolume  \undefined \def \bvolume#1{\textbf{#1}}\fi
\ifx \byear  \undefined \def \byear#1{#1}\fi
\ifx \bissue  \undefined \def \bissue#1{#1}\fi
\ifx \bfpage  \undefined \def \bfpage#1{#1}\fi
\ifx \blpage  \undefined \def \blpage #1{#1}\fi
\ifx \burl  \undefined \def \burl#1{\textsf{#1}}\fi
\ifx \doiurl  \undefined \def \doiurl#1{\url{https://doi.org/#1}}\fi
\ifx \betal  \undefined \def \betal{\textit{et al.}}\fi
\ifx \binstitute  \undefined \def \binstitute#1{#1}\fi
\ifx \binstitutionaled  \undefined \def \binstitutionaled#1{#1}\fi
\ifx \bctitle  \undefined \def \bctitle#1{#1}\fi
\ifx \beditor  \undefined \def \beditor#1{#1}\fi
\ifx \bpublisher  \undefined \def \bpublisher#1{#1}\fi
\ifx \bbtitle  \undefined \def \bbtitle#1{#1}\fi
\ifx \bedition  \undefined \def \bedition#1{#1}\fi
\ifx \bseriesno  \undefined \def \bseriesno#1{#1}\fi
\ifx \blocation  \undefined \def \blocation#1{#1}\fi
\ifx \bsertitle  \undefined \def \bsertitle#1{#1}\fi
\ifx \bsnm \undefined \def \bsnm#1{#1}\fi
\ifx \bsuffix \undefined \def \bsuffix#1{#1}\fi
\ifx \bparticle \undefined \def \bparticle#1{#1}\fi
\ifx \barticle \undefined \def \barticle#1{#1}\fi
\bibcommenthead
\ifx \bconfdate \undefined \def \bconfdate #1{#1}\fi
\ifx \botherref \undefined \def \botherref #1{#1}\fi
\ifx \url \undefined \def \url#1{\textsf{#1}}\fi
\ifx \bchapter \undefined \def \bchapter#1{#1}\fi
\ifx \bbook \undefined \def \bbook#1{#1}\fi
\ifx \bcomment \undefined \def \bcomment#1{#1}\fi
\ifx \oauthor \undefined \def \oauthor#1{#1}\fi
\ifx \citeauthoryear \undefined \def \citeauthoryear#1{#1}\fi
\ifx \endbibitem  \undefined \def \endbibitem {}\fi
\ifx \bconflocation  \undefined \def \bconflocation#1{#1}\fi
\ifx \arxivurl  \undefined \def \arxivurl#1{\textsf{#1}}\fi
\csname PreBibitemsHook\endcsname

\bibitem[\protect\citeauthoryear{He et~al.}{2023}]{he2023role}
\begin{barticle}
\bauthor{\bsnm{He}, \binits{X.}},
\bauthor{\bsnm{Khan}, \binits{S.}},
\bauthor{\bsnm{Ozturk}, \binits{I.}},
\bauthor{\bsnm{Murshed}, \binits{M.}}:
\batitle{The role of renewable energy investment in tackling climate change concerns: Environmental policies for achieving sdg-13}.
\bjtitle{Sustainable Development}
\bvolume{31}(\bissue{3}),
\bfpage{1888}--\blpage{1901}
(\byear{2023})
\end{barticle}
\endbibitem

\bibitem[\protect\citeauthoryear{Lorente et~al.}{2023}]{lorente2023dynamic}
\begin{barticle}
\bauthor{\bsnm{Lorente}, \binits{D.B.}},
\bauthor{\bsnm{Mohammed}, \binits{K.S.}},
\bauthor{\bsnm{Cifuentes-Faura}, \binits{J.}},
\bauthor{\bsnm{Shahzad}, \binits{U.}}:
\batitle{Dynamic connectedness among climate change index, green financial assets and renewable energy markets: Novel evidence from sustainable development perspective}.
\bjtitle{Renewable Energy}
\bvolume{204},
\bfpage{94}--\blpage{105}
(\byear{2023})
\end{barticle}
\endbibitem

\bibitem[\protect\citeauthoryear{Anser et~al.}{2024}]{anser2024formulating}
\begin{barticle}
\bauthor{\bsnm{Anser}, \binits{M.K.}},
\bauthor{\bsnm{Khan}, \binits{K.A.}},
\bauthor{\bsnm{Umar}, \binits{M.}},
\bauthor{\bsnm{Awosusi}, \binits{A.A.}},
\bauthor{\bsnm{Shamansurova}, \binits{Z.}}:
\batitle{Formulating sustainable development policy for a developed nation: exploring the role of renewable energy, natural gas efficiency and oil efficiency towards decarbonization}.
\bjtitle{International Journal of Sustainable Development \& World Ecology}
\bvolume{31}(\bissue{3}),
\bfpage{247}--\blpage{263}
(\byear{2024})
\end{barticle}
\endbibitem

\bibitem[\protect\citeauthoryear{Al-Ismail et~al.}{2023}]{al2023impacts}
\begin{barticle}
\bauthor{\bsnm{Al-Ismail}, \binits{F.S.}},
\bauthor{\bsnm{Alam}, \binits{M.S.}},
\bauthor{\bsnm{Shafiullah}, \binits{M.}},
\bauthor{\bsnm{Hossain}, \binits{M.I.}},
\bauthor{\bsnm{Rahman}, \binits{S.M.}}:
\batitle{Impacts of renewable energy generation on greenhouse gas emissions in saudi arabia: A comprehensive review}.
\bjtitle{Sustainability}
\bvolume{15}(\bissue{6}),
\bfpage{5069}
(\byear{2023})
\end{barticle}
\endbibitem

\bibitem[\protect\citeauthoryear{Yi et~al.}{2023}]{yi2023environmental}
\begin{barticle}
\bauthor{\bsnm{Yi}, \binits{S.}},
\bauthor{\bsnm{Abbasi}, \binits{K.R.}},
\bauthor{\bsnm{Hussain}, \binits{K.}},
\bauthor{\bsnm{Albaker}, \binits{A.}},
\bauthor{\bsnm{Alvarado}, \binits{R.}}:
\batitle{Environmental concerns in the united states: Can renewable energy, fossil fuel energy, and natural resources depletion help?}
\bjtitle{Gondwana Research}
\bvolume{117},
\bfpage{41}--\blpage{55}
(\byear{2023})
\end{barticle}
\endbibitem

\bibitem[\protect\citeauthoryear{Laskaratou}{2023}]{laskaratou2023energy}
\begin{botherref}
\oauthor{\bsnm{Laskaratou}, \binits{V.}}:
Energy dependence and national security.
Master's thesis,
$\Pi$$\alpha$$\nu$$\varepsilon$$\pi$$\iota$$\sigma$$\tau$$\acute{\eta}$$\mu$$\iota$o $\Pi$$\varepsilon$$\iota$$\rho$$\alpha$$\iota$$\acute{\omega}$$\varsigma$
(2023)
\end{botherref}
\endbibitem

\bibitem[\protect\citeauthoryear{Bamisile et~al.}{2023}]{bamisile2023geothermal}
\begin{barticle}
\bauthor{\bsnm{Bamisile}, \binits{O.}},
\bauthor{\bsnm{Cai}, \binits{D.}},
\bauthor{\bsnm{Adun}, \binits{H.}},
\bauthor{\bsnm{Taiwo}, \binits{M.}},
\bauthor{\bsnm{Li}, \binits{J.}},
\bauthor{\bsnm{Hu}, \binits{Y.}},
\bauthor{\bsnm{Huang}, \binits{Q.}}:
\batitle{Geothermal energy prospect for decarbonization, ewf nexus and energy poverty mitigation in east africa; the role of hydrogen production}.
\bjtitle{Energy Strategy Reviews}
\bvolume{49},
\bfpage{101157}
(\byear{2023})
\end{barticle}
\endbibitem

\bibitem[\protect\citeauthoryear{Qiu et~al.}{2023}]{qiu2023energy}
\begin{barticle}
\bauthor{\bsnm{Qiu}, \binits{L.}},
\bauthor{\bsnm{Wang}, \binits{X.}},
\bauthor{\bsnm{Wei}, \binits{J.}}:
\batitle{Energy security and energy management: The role of extreme natural events}.
\bjtitle{Innovation and Green Development}
\bvolume{2}(\bissue{2}),
\bfpage{100051}
(\byear{2023})
\end{barticle}
\endbibitem

\bibitem[\protect\citeauthoryear{Li et~al.}{2023}]{li2023energy}
\begin{barticle}
\bauthor{\bsnm{Li}, \binits{F.}},
\bauthor{\bsnm{Zhang}, \binits{J.}},
\bauthor{\bsnm{Li}, \binits{X.}}:
\batitle{Energy security dilemma and energy transition policy in the context of climate change: A perspective from china}.
\bjtitle{Energy Policy}
\bvolume{181},
\bfpage{113624}
(\byear{2023})
\end{barticle}
\endbibitem

\bibitem[\protect\citeauthoryear{Ang et~al.}{2022}]{ang2022comprehensive}
\begin{barticle}
\bauthor{\bsnm{Ang}, \binits{T.-Z.}},
\bauthor{\bsnm{Salem}, \binits{M.}},
\bauthor{\bsnm{Kamarol}, \binits{M.}},
\bauthor{\bsnm{Das}, \binits{H.S.}},
\bauthor{\bsnm{Nazari}, \binits{M.A.}},
\bauthor{\bsnm{Prabaharan}, \binits{N.}}:
\batitle{A comprehensive study of renewable energy sources: Classifications, challenges and suggestions}.
\bjtitle{Energy Strategy Reviews}
\bvolume{43},
\bfpage{100939}
(\byear{2022})
\end{barticle}
\endbibitem

\bibitem[\protect\citeauthoryear{Rizzi et~al.}{2014}]{rizzi2014production}
\begin{barticle}
\bauthor{\bsnm{Rizzi}, \binits{F.}},
\bauthor{\bsnm{Eck}, \binits{N.J.}},
\bauthor{\bsnm{Frey}, \binits{M.}}:
\batitle{The production of scientific knowledge on renewable energies: Worldwide trends, dynamics and challenges and implications for management}.
\bjtitle{Renewable Energy}
\bvolume{62},
\bfpage{657}--\blpage{671}
(\byear{2014})
\end{barticle}
\endbibitem

\bibitem[\protect\citeauthoryear{Ludin et~al.}{2018}]{ludin2018prospects}
\begin{barticle}
\bauthor{\bsnm{Ludin}, \binits{N.A.}},
\bauthor{\bsnm{Mustafa}, \binits{N.I.}},
\bauthor{\bsnm{Hanafiah}, \binits{M.M.}},
\bauthor{\bsnm{Ibrahim}, \binits{M.A.}},
\bauthor{\bsnm{Teridi}, \binits{M.A.M.}},
\bauthor{\bsnm{Sepeai}, \binits{S.}},
\bauthor{\bsnm{Zaharim}, \binits{A.}},
\bauthor{\bsnm{Sopian}, \binits{K.}}:
\batitle{Prospects of life cycle assessment of renewable energy from solar photovoltaic technologies: A review}.
\bjtitle{Renewable and Sustainable Energy Reviews}
\bvolume{96},
\bfpage{11}--\blpage{28}
(\byear{2018})
\end{barticle}
\endbibitem

\bibitem[\protect\citeauthoryear{{International Renewable Energy Agency (IRENA) and Renewable Energy Policy Network for the 21st Century (REN21)}}{2024}]{IRENA-REN21}
\begin{botherref}
\oauthor{\bsnm{{International Renewable Energy Agency (IRENA) and Renewable Energy Policy Network for the 21st Century (REN21)}}}:
{Global Renewable Energy Capacity by Source (2020-2023)}.
Data obtained from reports published by the International Renewable Energy Agency (IRENA) and the Renewable Energy Policy Network for the 21st Century (REN21).
(2024)
\end{botherref}
\endbibitem

\bibitem[\protect\citeauthoryear{Unknown}{2019}]{renewable_energy_ratio_2018}
\begin{botherref}
\oauthor{\bsnm{Unknown}}:
Renewable Energy Share in Electricity.
News.
Accessed on February 7, 2019
(2019)
\end{botherref}
\endbibitem

\bibitem[\protect\citeauthoryear{Bhuiyan et~al.}{2021}]{bhuiyan2021renewable}
\begin{barticle}
\bauthor{\bsnm{Bhuiyan}, \binits{M.A.}},
\bauthor{\bsnm{An}, \binits{J.}},
\bauthor{\bsnm{Mikhaylov}, \binits{A.}},
\bauthor{\bsnm{Moiseev}, \binits{N.}},
\bauthor{\bsnm{Danish}, \binits{M.S.S.}}:
\batitle{Renewable energy deployment and covid-19 measures for sustainable development}.
\bjtitle{Sustainability}
\bvolume{13}(\bissue{8}),
\bfpage{4418}
(\byear{2021})
\end{barticle}
\endbibitem

\bibitem[\protect\citeauthoryear{{BP}}{2020}]{statistical_review_world_energy_2020}
\begin{botherref}
\oauthor{\bsnm{{BP}}}:
Statistical Review of World Energy 2020.
\url{https://www.bp.com/content/dam/bp/business-sites/en/global/corporate/pdfs/energy-economics/statistical-review/bp-stats-review-2020-full-report.pdf}
Accessed 2021-04-15
\end{botherref}
\endbibitem

\bibitem[\protect\citeauthoryear{Petrovi{\'c} et~al.}{2020}]{petrovic2020importance}
\begin{barticle}
\bauthor{\bsnm{Petrovi{\'c}}, \binits{R.M.}},
\bauthor{\bsnm{Koci{\'c}}, \binits{N.}},
\bauthor{\bsnm{Stojanovi{\'c}}, \binits{R.B.}}:
\batitle{The importance of renewable energy sources for sustainable development}.
\bjtitle{Economics of Sustainable Development}
\bvolume{4}(\bissue{2}),
\bfpage{15}--\blpage{24}
(\byear{2020})
\end{barticle}
\endbibitem

\bibitem[\protect\citeauthoryear{Perea-Moreno}{2021}]{perea2021renewable}
\begin{botherref}
\oauthor{\bsnm{Perea-Moreno}, \binits{A.-J.}}:
Renewable energy and energy saving: Worldwide research trends.
MDPI
(2021)
\end{botherref}
\endbibitem

\bibitem[\protect\citeauthoryear{Alam and Murad}{2020}]{alam2020impacts}
\begin{barticle}
\bauthor{\bsnm{Alam}, \binits{M.M.}},
\bauthor{\bsnm{Murad}, \binits{M.W.}}:
\batitle{The impacts of economic growth, trade openness and technological progress on renewable energy use in organization for economic co-operation and development countries}.
\bjtitle{Renewable Energy}
\bvolume{145},
\bfpage{382}--\blpage{390}
(\byear{2020})
\end{barticle}
\endbibitem

\bibitem[\protect\citeauthoryear{Routledge}{2021}]{Routledge2021}
\begin{botherref}
\oauthor{\bsnm{Routledge}, \binits{T..F.J.}}:
Combating climate change: The role of science and engineering
(2021)
\end{botherref}
\endbibitem

\bibitem[\protect\citeauthoryear{Onwusinkwue et~al.}{2024}]{onwusinkwue2024artificial}
\begin{botherref}
\oauthor{\bsnm{Onwusinkwue}, \binits{S.}},
\oauthor{\bsnm{Osasona}, \binits{F.}},
\oauthor{\bsnm{Ahmad}, \binits{I.A.I.}},
\oauthor{\bsnm{Anyanwu}, \binits{A.C.}},
\oauthor{\bsnm{Dawodu}, \binits{S.O.}},
\oauthor{\bsnm{Obi}, \binits{O.C.}},
\oauthor{\bsnm{Hamdan}, \binits{A.}}:
Artificial intelligence (ai) in renewable energy: A review of predictive maintenance and energy optimization
(2024)
\end{botherref}
\endbibitem

\bibitem[\protect\citeauthoryear{Karim et~al.}{2017}]{karim2017electricity}
\begin{barticle}
\bauthor{\bsnm{Karim}, \binits{T.F.}},
\bauthor{\bsnm{Lipu}, \binits{M.H.}},
\bauthor{\bsnm{Mahmud}, \binits{M.S.}}:
\batitle{Electricity access improvement using renewable energy and energy efficiency: a case of urban poor area of dhaka, bangladesh}.
\bjtitle{International Journal of Renewable Energy Research (IJRER)}
\bvolume{7}(\bissue{3}),
\bfpage{1296}--\blpage{1306}
(\byear{2017})
\end{barticle}
\endbibitem

\bibitem[\protect\citeauthoryear{Tao et~al.}{2023}]{tao2023environmental}
\begin{barticle}
\bauthor{\bsnm{Tao}, \binits{Y.}},
\bauthor{\bsnm{Destek}, \binits{M.A.}},
\bauthor{\bsnm{Pata}, \binits{U.K.}},
\bauthor{\bsnm{Khan}, \binits{Z.}}:
\batitle{Environmental regulations and carbon emissions: the role of renewable energy research and development expenditures}.
\bjtitle{Sustainability}
\bvolume{15}(\bissue{18}),
\bfpage{13345}
(\byear{2023})
\end{barticle}
\endbibitem

\bibitem[\protect\citeauthoryear{Ahmad et~al.}{2021}]{ahmad2021artificial}
\begin{barticle}
\bauthor{\bsnm{Ahmad}, \binits{T.}},
\bauthor{\bsnm{Zhang}, \binits{D.}},
\bauthor{\bsnm{Huang}, \binits{C.}},
\bauthor{\bsnm{Zhang}, \binits{H.}},
\bauthor{\bsnm{Dai}, \binits{N.}},
\bauthor{\bsnm{Song}, \binits{Y.}},
\bauthor{\bsnm{Chen}, \binits{H.}}:
\batitle{Artificial intelligence in sustainable energy industry: Status quo, challenges and opportunities}.
\bjtitle{Journal of Cleaner Production}
\bvolume{289},
\bfpage{125834}
(\byear{2021})
\end{barticle}
\endbibitem

\bibitem[\protect\citeauthoryear{Georgiou}{2019}]{Georgiou2019}
\begin{botherref}
\oauthor{\bsnm{Georgiou}, \binits{M.}}:
The role of ai technology in improving the renewable energy sector.
Insider
(2019)
\end{botherref}
\endbibitem

\bibitem[\protect\citeauthoryear{Lahmadi}{2023}]{Lahmadi2023}
\begin{botherref}
\oauthor{\bsnm{Lahmadi}, \binits{K.}}:
Artificial intelligence can improve renewable energy production, report shows.
Morocco World News
(2023)
\end{botherref}
\endbibitem

\bibitem[\protect\citeauthoryear{Khan et~al.}{2021}]{khan2021db}
\begin{barticle}
\bauthor{\bsnm{Khan}, \binits{N.}},
\bauthor{\bsnm{Haq}, \binits{I.U.}},
\bauthor{\bsnm{Khan}, \binits{S.U.}},
\bauthor{\bsnm{Rho}, \binits{S.}},
\bauthor{\bsnm{Lee}, \binits{M.Y.}},
\bauthor{\bsnm{Baik}, \binits{S.W.}}:
\batitle{Db-net: A novel dilated cnn based multi-step forecasting model for power consumption in integrated local energy systems}.
\bjtitle{International Journal of Electrical Power \& Energy Systems}
\bvolume{133},
\bfpage{107023}
(\byear{2021})
\end{barticle}
\endbibitem

\bibitem[\protect\citeauthoryear{Nam et~al.}{2020}]{nam2020deep}
\begin{barticle}
\bauthor{\bsnm{Nam}, \binits{K.}},
\bauthor{\bsnm{Hwangbo}, \binits{S.}},
\bauthor{\bsnm{Yoo}, \binits{C.}}:
\batitle{A deep learning-based forecasting model for renewable energy scenarios to guide sustainable energy policy: A case study of korea}.
\bjtitle{Renewable and Sustainable Energy Reviews}
\bvolume{122},
\bfpage{109725}
(\byear{2020})
\end{barticle}
\endbibitem

\bibitem[\protect\citeauthoryear{Gensler et~al.}{2016}]{gensler2016deep}
\begin{bchapter}
\bauthor{\bsnm{Gensler}, \binits{A.}},
\bauthor{\bsnm{Henze}, \binits{J.}},
\bauthor{\bsnm{Sick}, \binits{B.}},
\bauthor{\bsnm{Raabe}, \binits{N.}}:
\bctitle{Deep learning for solar power forecasting—an approach using autoencoder and lstm neural networks}.
In: \bbtitle{2016 IEEE International Conference on Systems, Man, and Cybernetics (SMC)},
pp. \bfpage{002858}--\blpage{002865}
(\byear{2016}).
\bcomment{IEEE}
\end{bchapter}
\endbibitem

\bibitem[\protect\citeauthoryear{Khan et~al.}{2021}]{khan2021ab}
\begin{barticle}
\bauthor{\bsnm{Khan}, \binits{N.}},
\bauthor{\bsnm{Ullah}, \binits{F.U.M.}},
\bauthor{\bsnm{Haq}, \binits{I.U.}},
\bauthor{\bsnm{Khan}, \binits{S.U.}},
\bauthor{\bsnm{Lee}, \binits{M.Y.}},
\bauthor{\bsnm{Baik}, \binits{S.W.}}:
\batitle{Ab-net: A novel deep learning assisted framework for renewable energy generation forecasting}.
\bjtitle{Mathematics}
\bvolume{9}(\bissue{19}),
\bfpage{2456}
(\byear{2021})
\end{barticle}
\endbibitem

\bibitem[\protect\citeauthoryear{Mostafa et~al.}{2022}]{mostafa2022renewable}
\begin{barticle}
\bauthor{\bsnm{Mostafa}, \binits{N.}},
\bauthor{\bsnm{Ramadan}, \binits{H.S.M.}},
\bauthor{\bsnm{Elfarouk}, \binits{O.}}:
\batitle{Renewable energy management in smart grids by using big data analytics and machine learning}.
\bjtitle{Machine Learning with Applications}
\bvolume{9},
\bfpage{100363}
(\byear{2022})
\end{barticle}
\endbibitem

\bibitem[\protect\citeauthoryear{Gr{\`e}ve et~al.}{2020}]{greve2020machine}
\begin{barticle}
\bauthor{\bsnm{Gr{\`e}ve}, \binits{Z.D.}},
\bauthor{\bsnm{Bottieau}, \binits{J.}},
\bauthor{\bsnm{Vangulick}, \binits{D.}},
\bauthor{\bsnm{Wautier}, \binits{A.}},
\bauthor{\bsnm{Dapoz}, \binits{P.-D.}},
\bauthor{\bsnm{Arrigo}, \binits{A.}},
\bauthor{\bsnm{Toubeau}, \binits{J.-F.}},
\bauthor{\bsnm{Vall{\'e}e}, \binits{F.}}:
\batitle{Machine learning techniques for improving self-consumption in renewable energy communities}.
\bjtitle{Energies}
\bvolume{13}(\bissue{18}),
\bfpage{4892}
(\byear{2020})
\end{barticle}
\endbibitem

\bibitem[\protect\citeauthoryear{Solano et~al.}{2022}]{solano2022solar}
\begin{barticle}
\bauthor{\bsnm{Solano}, \binits{E.S.}},
\bauthor{\bsnm{Dehghanian}, \binits{P.}},
\bauthor{\bsnm{Affonso}, \binits{C.M.}}:
\batitle{Solar radiation forecasting using machine learning and ensemble feature selection}.
\bjtitle{Energies}
\bvolume{15}(\bissue{19}),
\bfpage{7049}
(\byear{2022})
\end{barticle}
\endbibitem

\bibitem[\protect\citeauthoryear{Conte et~al.}{2022}]{conte2022new}
\begin{barticle}
\bauthor{\bsnm{Conte}, \binits{F.}},
\bauthor{\bsnm{D’Antoni}, \binits{F.}},
\bauthor{\bsnm{Natrella}, \binits{G.}},
\bauthor{\bsnm{Merone}, \binits{M.}}:
\batitle{A new hybrid ai optimal management method for renewable energy communities}.
\bjtitle{Energy and AI}
\bvolume{10},
\bfpage{100197}
(\byear{2022})
\end{barticle}
\endbibitem

\bibitem[\protect\citeauthoryear{Azhar et~al.}{2022}]{azhar2022development}
\begin{barticle}
\bauthor{\bsnm{Azhar}, \binits{I.F.}},
\bauthor{\bsnm{Putranto}, \binits{L.M.}},
\bauthor{\bsnm{Irnawan}, \binits{R.}}:
\batitle{Development of pmu-based transient stability detection methods using cnn-lstm considering time series data measurement}.
\bjtitle{Energies}
\bvolume{15}(\bissue{21}),
\bfpage{8241}
(\byear{2022})
\end{barticle}
\endbibitem

\bibitem[\protect\citeauthoryear{Sani et~al.}{2019}]{sani2019cyber}
\begin{barticle}
\bauthor{\bsnm{Sani}, \binits{A.S.}},
\bauthor{\bsnm{Yuan}, \binits{D.}},
\bauthor{\bsnm{Jin}, \binits{J.}},
\bauthor{\bsnm{Gao}, \binits{L.}},
\bauthor{\bsnm{Yu}, \binits{S.}},
\bauthor{\bsnm{Dong}, \binits{Z.Y.}}:
\batitle{Cyber security framework for internet of things-based energy internet}.
\bjtitle{Future Generation Computer Systems}
\bvolume{93},
\bfpage{849}--\blpage{859}
(\byear{2019})
\end{barticle}
\endbibitem

\bibitem[\protect\citeauthoryear{Vernotte et~al.}{2018}]{vernotte2018load}
\begin{barticle}
\bauthor{\bsnm{Vernotte}, \binits{A.}},
\bauthor{\bsnm{V{\"a}lja}, \binits{M.}},
\bauthor{\bsnm{Korman}, \binits{M.}},
\bauthor{\bsnm{Bj{\"o}rkman}, \binits{G.}},
\bauthor{\bsnm{Ekstedt}, \binits{M.}},
\bauthor{\bsnm{Lagerstr{\"o}m}, \binits{R.}}:
\batitle{Load balancing of renewable energy: a cyber security analysis}.
\bjtitle{Energy Informatics}
\bvolume{1},
\bfpage{1}--\blpage{41}
(\byear{2018})
\end{barticle}
\endbibitem

\bibitem[\protect\citeauthoryear{Mazzeo et~al.}{2021}]{mazzeo2021artificial}
\begin{barticle}
\bauthor{\bsnm{Mazzeo}, \binits{D.}},
\bauthor{\bsnm{Herdem}, \binits{M.S.}},
\bauthor{\bsnm{Matera}, \binits{N.}},
\bauthor{\bsnm{Bonini}, \binits{M.}},
\bauthor{\bsnm{Wen}, \binits{J.Z.}},
\bauthor{\bsnm{Nathwani}, \binits{J.}},
\bauthor{\bsnm{Oliveti}, \binits{G.}}:
\batitle{Artificial intelligence application for the performance prediction of a clean energy community}.
\bjtitle{Energy}
\bvolume{232},
\bfpage{120999}
(\byear{2021})
\end{barticle}
\endbibitem

\bibitem[\protect\citeauthoryear{Rekeraho et~al.}{2024}]{rekeraho2024cybersecurity}
\begin{barticle}
\bauthor{\bsnm{Rekeraho}, \binits{A.}},
\bauthor{\bsnm{Cotfas}, \binits{D.T.}},
\bauthor{\bsnm{Cotfas}, \binits{P.A.}},
\bauthor{\bsnm{B{\u{a}}lan}, \binits{T.C.}},
\bauthor{\bsnm{Tuyishime}, \binits{E.}},
\bauthor{\bsnm{Acheampong}, \binits{R.}}:
\batitle{Cybersecurity challenges in iot-based smart renewable energy}.
\bjtitle{International Journal of Information Security}
\bvolume{23}(\bissue{1}),
\bfpage{101}--\blpage{117}
(\byear{2024})
\end{barticle}
\endbibitem

\bibitem[\protect\citeauthoryear{Ghiasi et~al.}{2023}]{ghiasi2023comprehensive}
\begin{barticle}
\bauthor{\bsnm{Ghiasi}, \binits{M.}},
\bauthor{\bsnm{Niknam}, \binits{T.}},
\bauthor{\bsnm{Wang}, \binits{Z.}},
\bauthor{\bsnm{Mehrandezh}, \binits{M.}},
\bauthor{\bsnm{Dehghani}, \binits{M.}},
\bauthor{\bsnm{Ghadimi}, \binits{N.}}:
\batitle{A comprehensive review of cyber-attacks and defense mechanisms for improving security in smart grid energy systems: Past, present and future}.
\bjtitle{Electric Power Systems Research}
\bvolume{215},
\bfpage{108975}
(\byear{2023})
\end{barticle}
\endbibitem

\bibitem[\protect\citeauthoryear{Ahmadi et~al.}{2022}]{ahmadi2022new}
\begin{barticle}
\bauthor{\bsnm{Ahmadi}, \binits{A.}},
\bauthor{\bsnm{Nabipour}, \binits{M.}},
\bauthor{\bsnm{Taheri}, \binits{S.}},
\bauthor{\bsnm{Mohammadi-Ivatloo}, \binits{B.}},
\bauthor{\bsnm{Vahidinasab}, \binits{V.}}:
\batitle{A new false data injection attack detection model for cyberattack resilient energy forecasting}.
\bjtitle{IEEE Transactions on Industrial Informatics}
\bvolume{19}(\bissue{1}),
\bfpage{371}--\blpage{381}
(\byear{2022})
\end{barticle}
\endbibitem

\bibitem[\protect\citeauthoryear{Barreto et~al.}{2020}]{barreto2020cyber}
\begin{bchapter}
\bauthor{\bsnm{Barreto}, \binits{C.}},
\bauthor{\bsnm{Eghtesad}, \binits{T.}},
\bauthor{\bsnm{Eisele}, \binits{S.}},
\bauthor{\bsnm{Laszka}, \binits{A.}},
\bauthor{\bsnm{Dubey}, \binits{A.}},
\bauthor{\bsnm{Koutsoukos}, \binits{X.}}:
\bctitle{Cyber-attacks and mitigation in blockchain based transactive energy systems}.
In: \bbtitle{2020 IEEE Conference on Industrial Cyberphysical Systems (ICPS)},
vol. \bseriesno{1},
pp. \bfpage{129}--\blpage{136}
(\byear{2020}).
\bcomment{IEEE}
\end{bchapter}
\endbibitem

\bibitem[\protect\citeauthoryear{Ilbeigi et~al.}{2020}]{ilbeigi2020prediction}
\begin{barticle}
\bauthor{\bsnm{Ilbeigi}, \binits{M.}},
\bauthor{\bsnm{Ghomeishi}, \binits{M.}},
\bauthor{\bsnm{Dehghanbanadaki}, \binits{A.}}:
\batitle{Prediction and optimization of energy consumption in an office building using artificial neural network and a genetic algorithm}.
\bjtitle{Sustainable Cities and Society}
\bvolume{61},
\bfpage{102325}
(\byear{2020})
\end{barticle}
\endbibitem

\bibitem[\protect\citeauthoryear{Guo et~al.}{2019}]{guo2019expert}
\begin{barticle}
\bauthor{\bsnm{Guo}, \binits{Y.}},
\bauthor{\bsnm{Wang}, \binits{J.}},
\bauthor{\bsnm{Chen}, \binits{H.}},
\bauthor{\bsnm{Li}, \binits{G.}},
\bauthor{\bsnm{Huang}, \binits{R.}},
\bauthor{\bsnm{Yuan}, \binits{Y.}},
\bauthor{\bsnm{Ahmad}, \binits{T.}},
\bauthor{\bsnm{Sun}, \binits{S.}}:
\batitle{An expert rule-based fault diagnosis strategy for variable refrigerant flow air conditioning systems}.
\bjtitle{Applied Thermal Engineering}
\bvolume{149},
\bfpage{1223}--\blpage{1235}
(\byear{2019})
\doiurl{10.1016/j.applthermaleng.2018.12.132}
\end{barticle}
\endbibitem

\bibitem[\protect\citeauthoryear{Zhao et~al.}{2013}]{zhao2013intelligent}
\begin{barticle}
\bauthor{\bsnm{Zhao}, \binits{Y.}},
\bauthor{\bsnm{Xiao}, \binits{F.}},
\bauthor{\bsnm{Wang}, \binits{S.}}:
\batitle{An intelligent chiller fault detection and diagnosis methodology using bayesian belief network}.
\bjtitle{Energy and Buildings}
\bvolume{57},
\bfpage{278}--\blpage{288}
(\byear{2013})
\doiurl{10.1016/j.enbuild.2012.11.007}
\end{barticle}
\endbibitem

\bibitem[\protect\citeauthoryear{Tang et~al.}{2018}]{tang2018framework}
\begin{barticle}
\bauthor{\bsnm{Tang}, \binits{Y.}},
\bauthor{\bsnm{Huang}, \binits{Y.}},
\bauthor{\bsnm{Wang}, \binits{H.}},
\bauthor{\bsnm{Wang}, \binits{C.}},
\bauthor{\bsnm{Guo}, \binits{Q.}},
\bauthor{\bsnm{Yao}, \binits{W.}}:
\batitle{Framework for artificial intelligence analysis in large-scale power grids based on digital simulation}.
\bjtitle{CSEE Journal of Power and Energy Systems}
\bvolume{4},
\bfpage{459}--\blpage{468}
(\byear{2018})
\doiurl{10.17775/cseejpes.2018.01010}
\end{barticle}
\endbibitem

\bibitem[\protect\citeauthoryear{Puri et~al.}{2019}]{puri2019hybrid}
\begin{barticle}
\bauthor{\bsnm{Puri}, \binits{V.}},
\bauthor{\bsnm{Jha}, \binits{S.}},
\bauthor{\bsnm{Kumar}, \binits{R.}},
\bauthor{\bsnm{Priyadarshini}, \binits{I.}},
\bauthor{\bsnm{Hoang~Son}, \binits{L.}},
\bauthor{\bsnm{Abdel-Basset}, \binits{M.}},
\bauthor{\bsnm{Elhoseny}, \binits{M.}},
\bauthor{\bsnm{Viet~Long}, \binits{H.}}:
\batitle{A hybrid artificial intelligence and internet of things model for generation of renewable resource of energy}.
\bjtitle{IEEE Access}
\bvolume{7},
\bfpage{111181}--\blpage{111191}
(\byear{2019})
\doiurl{10.1109/access.2019.2934228}
\end{barticle}
\endbibitem

\bibitem[\protect\citeauthoryear{Al~Nasim et~al.}{2022}]{10054934}
\begin{bchapter}
\bauthor{\bsnm{Al~Nasim}, \binits{M.A.}},
\bauthor{\bsnm{Al~Munem}, \binits{A.}},
\bauthor{\bsnm{Islam}, \binits{M.}},
\bauthor{\bsnm{Palash}, \binits{M.A.H.}},
\bauthor{\bsnm{Haque}, \binits{M.M.A.}},
\bauthor{\bsnm{Shah}, \binits{F.M.}}:
\bctitle{Brain tumor segmentation using enhanced u-net model with empirical analysis}.
In: \bbtitle{2022 25th International Conference on Computer and Information Technology (ICCIT)},
pp. \bfpage{1027}--\blpage{1032}
(\byear{2022}).
\doiurl{10.1109/ICCIT57492.2022.10054934}
\end{bchapter}
\endbibitem

\bibitem[\protect\citeauthoryear{Biswas and Islam}{2023}]{biswas2023hybrid}
\begin{botherref}
\oauthor{\bsnm{Biswas}, \binits{A.}},
\oauthor{\bsnm{Islam}, \binits{M.S.}}:
A hybrid deep cnn-svm approach for brain tumor classification.
Journal of Information Systems Engineering \& Business Intelligence
\textbf{9}(1)
(2023)
\end{botherref}
\endbibitem

\bibitem[\protect\citeauthoryear{Biswas and Islam}{2022}]{biswas2022mri}
\begin{bchapter}
\bauthor{\bsnm{Biswas}, \binits{A.}},
\bauthor{\bsnm{Islam}, \binits{M.S.}}:
\bctitle{Mri brain tumor classification technique using fuzzy c-means clustering and artificial neural network}.
In: \beditor{\bsnm{Ibrahim}, \binits{R.}},
\beditor{\bsnm{Porkumaran}, \binits{K.}},
\beditor{\bsnm{Kannan}, \binits{R.}},
\beditor{\bsnm{Mohd~Nor}, \binits{N.}},
\beditor{\bsnm{Prabakar}, \binits{S.}} (eds.)
\bbtitle{International Conference on Artificial Intelligence for Smart Community}.
\bsertitle{Lecture Notes in Electrical Engineering},
vol. \bseriesno{758}.
\bpublisher{Springer}, \blocation{???}
(\byear{2022}).
\doiurl{10.1007/978-981-16-2183-3_95} .
\burl{https://doi.org/10.1007/978-981-16-2183-3_95}
\end{bchapter}
\endbibitem

\bibitem[\protect\citeauthoryear{Biswas et~al.}{2023a}]{biswas2023active}
\begin{bchapter}
\bauthor{\bsnm{Biswas}, \binits{A.}},
\bauthor{\bsnm{Abdullah~Al}, \binits{N.M.}},
\bauthor{\bsnm{Ali}, \binits{M.S.}},
\bauthor{\bsnm{Hossain}, \binits{I.}},
\bauthor{\bsnm{Ullah}, \binits{M.A.}},
\bauthor{\bsnm{Talukder}, \binits{S.}}:
\bctitle{Active learning on medical image}.
In: \bbtitle{Data Driven Approaches on Medical Imaging},
pp. \bfpage{51}--\blpage{67}.
\bpublisher{Springer}, \blocation{???}
(\byear{2023})
\end{bchapter}
\endbibitem

\bibitem[\protect\citeauthoryear{Biswas et~al.}{2023b}]{biswas2023generative}
\begin{bchapter}
\bauthor{\bsnm{Biswas}, \binits{A.}},
\bauthor{\bsnm{Md~Abdullah~Al}, \binits{N.}},
\bauthor{\bsnm{Imran}, \binits{A.}},
\bauthor{\bsnm{Sejuty}, \binits{A.T.}},
\bauthor{\bsnm{Fairooz}, \binits{F.}},
\bauthor{\bsnm{Puppala}, \binits{S.}},
\bauthor{\bsnm{Talukder}, \binits{S.}}:
\bctitle{Generative adversarial networks for data augmentation}.
In: \bbtitle{Data Driven Approaches on Medical Imaging},
pp. \bfpage{159}--\blpage{177}.
\bpublisher{Springer}, \blocation{???}
(\byear{2023})
\end{bchapter}
\endbibitem

\bibitem[\protect\citeauthoryear{Irbaz et~al.}{2022}]{irbaz2022real}
\begin{bchapter}
\bauthor{\bsnm{Irbaz}, \binits{M.S.}},
\bauthor{\bsnm{Al~Nasim}, \binits{M.A.}},
\bauthor{\bsnm{Ferdous}, \binits{R.E.}}:
\bctitle{Real-time face recognition system for remote employee tracking}.
In: \beditor{\bsnm{Arefin}, \binits{M.S.}},
\beditor{\bsnm{Kaiser}, \binits{M.S.}},
\beditor{\bsnm{Bandyopadhyay}, \binits{A.}},
\beditor{\bsnm{Ahad}, \binits{M.A.R.}},
\beditor{\bsnm{Ray}, \binits{K.}} (eds.)
\bbtitle{Proceedings of the International Conference on Big Data, IoT, and Machine Learning}.
\bsertitle{Lecture Notes on Data Engineering and Communications Technologies},
vol. \bseriesno{95}.
\bpublisher{Springer}, \blocation{???}
(\byear{2022}).
\doiurl{10.1007/978-981-16-6636-0_13} .
\burl{https://doi.org/10.1007/978-981-16-6636-0_13}
\end{bchapter}
\endbibitem

\bibitem[\protect\citeauthoryear{Biswas and Islam}{2021}]{biswas2021efficient}
\begin{barticle}
\bauthor{\bsnm{Biswas}, \binits{A.}},
\bauthor{\bsnm{Islam}, \binits{M.S.}}:
\batitle{An efficient cnn model for automated digital handwritten digit classification}.
\bjtitle{Journal of Information Systems Engineering and Business Intelligence}
\bvolume{7}(\bissue{1}),
\bfpage{42}--\blpage{55}
(\byear{2021})
\end{barticle}
\endbibitem

\bibitem[\protect\citeauthoryear{}{}]{datagokr}
\begin{botherref}
{DATA.GO.KR}.
Accessed on 5 April 2021
\end{botherref}
\endbibitem

\bibitem[\protect\citeauthoryear{}{}]{nrelwind}
\begin{botherref}
{NREL Wind Prospector}.
Accessed on 5 April 2021
\end{botherref}
\endbibitem

\bibitem[\protect\citeauthoryear{Arzamasov et~al.}{2018}]{8587498}
\begin{bchapter}
\bauthor{\bsnm{Arzamasov}, \binits{V.}},
\bauthor{\bsnm{Böhm}, \binits{K.}},
\bauthor{\bsnm{Jochem}, \binits{P.}}:
\bctitle{Towards concise models of grid stability}.
In: \bbtitle{2018 IEEE International Conference on Communications, Control, and Computing Technologies for Smart Grids (SmartGridComm)},
pp. \bfpage{1}--\blpage{6}
(\byear{2018}).
\doiurl{10.1109/SmartGridComm.2018.8587498}
\end{bchapter}
\endbibitem

\bibitem[\protect\citeauthoryear{Shen et~al.}{2018}]{shen2018combined}
\begin{barticle}
\bauthor{\bsnm{Shen}, \binits{X.}},
\bauthor{\bsnm{Fu}, \binits{X.}},
\bauthor{\bsnm{Zhou}, \binits{C.}}:
\batitle{A combined algorithm for cleaning abnormal data of wind turbine power curve based on change point grouping algorithm and quartile algorithm}.
\bjtitle{IEEE Transactions on Sustainable Energy}
\bvolume{10}(\bissue{1}),
\bfpage{46}--\blpage{54}
(\byear{2018})
\end{barticle}
\endbibitem

\bibitem[\protect\citeauthoryear{}{}]{inmet}
\begin{botherref}
{INMET. Instituto Nacional de Meteorologia}.
Accessed on 23 October 2021.
\url{https://portal.inmet.gov.br/}
Accessed 2021-10-23
\end{botherref}
\endbibitem

\bibitem[\protect\citeauthoryear{Ramos et~al.}{2021}]{ramos_energy_2021}
\begin{botherref}
\oauthor{\bsnm{Ramos}, \binits{S.}},
\oauthor{\bsnm{Soares}, \binits{J.}},
\oauthor{\bsnm{Foroozandeh}, \binits{Z.}},
\oauthor{\bsnm{Tavares}, \binits{I.}},
\oauthor{\bsnm{Vale}, \binits{Z.}}:
Energy consumption and PV generation data of 15 prosumers (15 minute resolution).
Zenodo.
Accessed on [date]
(2021).
\url{http://dx.doi.org/10.5281/zenodo.5106455}
\end{botherref}
\endbibitem

\bibitem[\protect\citeauthoryear{Izzuddinfa}{year}]{izzuddinfa_transient_stability}
\begin{botherref}
\oauthor{\bsnm{Izzuddinfa}}:
Transient Stability.
Kaggle.
\url{https://www.kaggle.com/datasets/izzuddinfa/transient-stability}
(year)
\end{botherref}
\endbibitem

\end{thebibliography}

\end{document}